\pdfoutput=1
\documentclass[10pt,twocolumn,letterpaper]{article}
\usepackage[pagenumbers]{cvpr}
\usepackage{amsmath,amssymb,amsthm,mathtools}
\usepackage{booktabs,array,multirow}
\usepackage{graphicx,microtype,enumitem,needspace,float}
\definecolor{cvprblue}{rgb}{0.21,0.49,0.74}
\usepackage[pagebackref,breaklinks,colorlinks,allcolors=cvprblue]{hyperref}
\usepackage{bookmark}
\makeatletter
\def\NAT@sort{0}\def\NAT@cmprs{0}
\AtBeginDocument{%
  \let\CAM@BR@starttoc\BR@starttoc
  \def\BR@starttoc{%
    \CAM@BR@starttoc
    \global\let\BR@starttoc\relax
  }%
}
\makeatother
\newtheorem{theorem}{Theorem}

\theoremstyle{definition}

\setlist[itemize]{leftmargin=*,nosep}
\def\confName{CVPR}\def\confYear{2027}
\title{Minimal Memory for Correctable Visual Continuation}
\author{
  Xiyang Zhang\textsuperscript{1,2}\quad
  Yuanhe Tian\textsuperscript{2,3}\thanks{Corresponding author}\quad
  Hongzhi Wang\textsuperscript{1}\quad
  Yan Song\textsuperscript{4}\\[0.5em]
  {\normalfont
  \textsuperscript{1}Harbin Institute of Technology,\quad
  \textsuperscript{2}Zhongguancun Academy}\\
  {\normalfont
  \textsuperscript{3}Zhongguancun Institute of Artificial Intelligence}\\
  {\normalfont
  \textsuperscript{4}University of Science and Technology of China}\\[0.5em]
  {\normalfont\small
  \texttt{s-zhangxiy24@bza.edu.cn}\quad
  \texttt{tianyuanhe@zgci.ac.cn}}\\
  {\normalfont\small
  \texttt{wangzh@hit.edu.cn}\quad
  \texttt{clksong@gmail.com}}
}
\BeforeBeginEnvironment{theorem}{\Needspace{5\baselineskip}}
\BeforeBeginEnvironment{lemma}{\Needspace{5\baselineskip}}
\BeforeBeginEnvironment{proposition}{\Needspace{5\baselineskip}}

\begin{document}
\lefthyphenmin=2 \righthyphenmin=3
\maketitle

\begin{abstract}
Given past images and known actions, visual continuation generates future frames from one evolving scene state. Keeping this state revisable matters when an object reappears or a later view changes an earlier interpretation. Once past images cannot be replayed, a summary sufficient for current rendering can lose candidates needed after correction. We establish matching sparse and dense bounds on the bits needed for later correction under stable observations and bounded disagreement. For static streams at the stated smoothness and margin scales, correction incurs at least logarithmic overhead in one dimension and polynomial overhead in higher dimensions. Coherent Archive Memory (CAM) stores geometric summaries with exact updates under clipped expansion and coordinate assignments, then uses historical existence constraints to select one state for all requested frames. Exhaustive photographic tests show strict worst-truth image-risk gains from joint witnesses, while controlled neural continuations recover positions hidden in the current view.
\end{abstract}

\section{Introduction}
Long visual narratives hide and reveal objects as actions change their states. Given earlier images, known actions, a new observation and a future frame request, visual continuation generates a sequence explained by one evolving state. A mug moved behind a screen should reappear at its new location; a later view may revise an earlier belief. Coherence requires preserving action consequences while revising uncertain interpretations across shots and revisits.

Existing systems already condense visual history. StoryMem retains compact keyframes~\cite{zhang2025storymem}, while fixed-capacity spatial memories evict observations~\cite{alradi2026keepsake}. World models update retained evidence under actions~\cite{xiao2025worldmem,wu2025spatialmemory,garcin2026persist,lillemark2026flowm,venturoli2026statecraft,ki2026stateworld}. Timed reference uploads and dynamic-memory benchmarks expose changing evidence~\cite{xu2026reel,zhou2026worldmembench}. Once an observation cannot be replayed, queries must rely on the distinctions retained in memory.

A candidate can look dispensable before distinguishing evidence arrives. Two histories may have the same rendering decision while only one contains an intermediate explanation. A later view ruling out an old extreme can make that explanation determine the continuation. Preserving only the old decision loses this distinction. Separate attribute summaries can also lose which values belonged together. If an apple reveals the location of an occluded cup through their earlier relation, an exchange action transfers that relation to the other object.

We ask what history must survive before the correcting view is known. At a known state--renderer interface, past images supply candidate explanations, actions transport them to the present, and later evidence filters them. One feasible state generates the requested block. The full-history oracle minimizes worst rendered disagreement with all surviving candidates; compression excess measures the additional loss due to memory. This separates missing information from ambiguity already present in the history. Candidate coverage and renderer calibration connect this decision to physical prediction.

Our main result identifies how dimension changes the information needed for correction. We establish matching scalar and vector memory laws for sparse histories and dense archives, including randomized expected excess. Dense vector memory must preserve the shape inside each occupied region; two scalar extrema no longer suffice. For static streams, correction at the stated scales incurs at least logarithmic overhead in one dimension and polynomial overhead in higher dimensions. A finite bank of quadratic profiles attains the vector bound through exact integer updates, even when actions copy, exchange, reflect or erase attributes. These assignments can occur arbitrarily often; only expanding actions consume the horizon budget. Historical witnesses then do more than tighten an excess certificate: together with the given disagreement bound, they identify which true states are consistent with the retained constraints.

These information limits hold even with unrestricted readout computation.

CAM first encodes the joint candidate geometry, then updates it under actions, conditions it on later evidence and selects one state for the future frames. Each stored geometric measurement also records that some historical candidate existed in a known region. Combining these existence constraints with a disagreement bound can rule out states that a containing pool alone would retain. Photographic tests demonstrate joint-witness risk reduction, and neural continuations recover occluded positions.

\Needspace{8\baselineskip}
Our contributions are:
\begin{itemize}
\item We prove matching sparse and dense vector memory laws and show that, for static streams at the stated scales, correction incurs at least logarithmic overhead in one dimension and polynomial overhead in higher dimensions.
\item We construct finite-bit geometric summaries whose statistics update exactly under the stated actions, without repeated quantization of past candidates.
\item We use historical existence constraints and a given disagreement bound to exclude unsupported candidate states and select a common continuation state with minimum worst-case physical risk.
\end{itemize}

\section{Related Work}

\noindent\textbf{Visual memory and state evolution.} Visual references, spatial state and retrieval support continuations~\cite{rahman2023makeastory,tewel2024consistory,zhou2024storydiffusion,xiao2025worldmem,wu2025spatialmemory,garcin2026persist,lillemark2026flowm,wu2026worldtrace}. Valid-time memory uses action logs to date retained states~\cite{wang2026redate}. Evolving predictors and agent harnesses reuse experience under changing objectives~\cite{zhang2026worldevolver,yu2026recuris,cui2026jepaanything,zeng2026memowm,cui2026rsi}. CAM counts the bits needed before the future view or objective is known.

\noindent\textbf{Geometric summaries.} Entropy, convex approximation and streaming coresets supply approximation tools~\cite{clements1963entropies,dudley1974entropy,bronshtein1976entropy,agarwal2004extent,chan2006coresets,agarwal2007spaceoptimal,yu2008kinetic,chan2009dynamiccoresets,guntuboyina2013covering,chan2016dynamic,bishnu2018streaming,nasser2020kinematic}. Loss-independent and instance-optimal kernels address unknown objectives or point counts~\cite{lu2020robustcoreset,agarwal2023instance}. Range-oracle hull recovery instead charges query access~\cite{schibler2026rangehull}. We charge stored bits under chronological actions and later correction.

\noindent\textbf{State abstraction and correction.} State abstraction formalizes future-relevant distinctions~\cite{nerode1958automaton,giacobazzi2000complete,littman2001predictive,girard2007metrics,pola2008symbolic,mahony2022equivariance,jiao2026observable,zou2026decisionmemory,li2026recurrentmemory}. Set-membership estimation propagates sound feasible states~\cite{bertsekas1971recursive,milanese1991optimal}. We characterize the additional information needed when observations arrive after irreversible compression.

\noindent\textbf{Information and conditional decisions.} Support reconstruction and minimax regret motivate exact readout~\cite{prince1990support,averbakh2000minmax,averbakh2005facility,frehse2013flowpipe,mastin2015randomized,frehse2015eliminating,cai2018planar}. Universal clustering fixes its metric before clients arrive~\cite{ganesh2021universal}. Range-emptiness and augmented-index bounds study retained membership information~\cite{chakrabarti2013augmented,goswami2015range}. Here actions and later observations determine the retained information and output accuracy.

\section{The Continuation Problem}
\label{sec:model}
\begin{figure*}[t]
\centering\includegraphics[width=\textwidth]{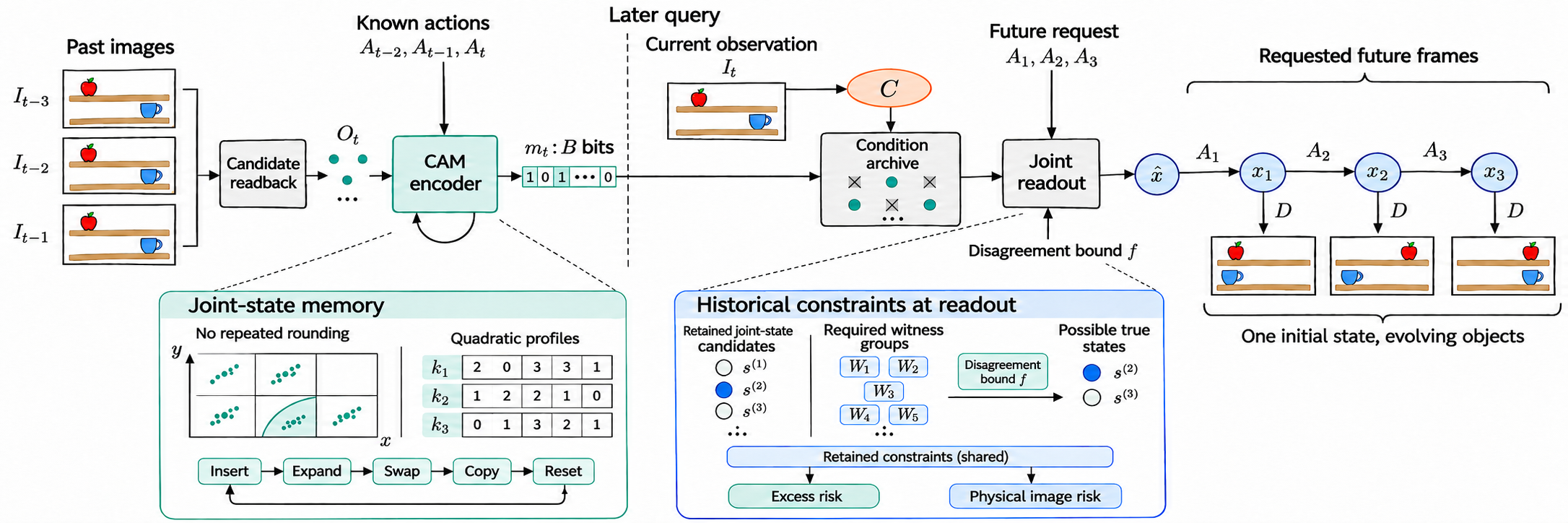}
\caption{\textbf{Coherent Archive Memory (CAM).} A $B$-bit packet transports joint geometry and witnesses exactly. Later evidence and a disagreement bound $f$ constrain one state for all future frames. Thumbnails and profiles are schematic.}
\label{fig:method}
\end{figure*}

We formalize the interface in Fig.~\ref{fig:method}, beginning with the reference decision available with complete history.

\subsection{Candidates, correction, and a common output state}
A scene state $x\in X$ specifies the variables needed for rendering, with compact domain $X$. A past image supplies a finite set $O_t\subset X$ of candidate explanations, such as an object phase or a pair of horizontal object positions. A missing measurement supplies no candidates. Because an occluded or mistaken readback can reject a useful explanation, the archive defers rejection to the later correcting view. A known action $A_t:X\to X$ transports earlier proposals into present coordinates:
\begin{equation}
V_t=A_t(V_{t-1})\cup O_t.
\label{eq:chronological}
\end{equation}
Thus $V_t$ is an archive of transported proposals, whose members represent alternative worlds. A later observation supplies a closed set $C\subseteq X$ and rejects proposals outside it. Write $V_C=V_t\cap C$ and assume $V_C\ne\varnothing$.

The requested sequence specifies maps $G_j:X\to[0,1]^{P_j}$, $j=1,\ldots,T$. Each $G_j$ includes the requested future actions and its image renderer. Choosing one current state $x\in C$ produces all frames $G_1(x),\ldots,G_T(x)$; separate states cannot be chosen independently for different frames. With normalized image norm $\|z\|_P=\|z\|_2/\sqrt P$, define
\begin{align}
\ell_G(u,x)&=\max_{1\le j\le T}\|G_j(u)-G_j(x)\|_{P_j}^2,\nonumber\\
L^C(V,x)&=\max_{u\in V_C}\ell_G(u,x).
\label{eq:conditional-loss}
\end{align}
The first maximum selects the worst requested frame. The second selects the surviving historical candidate with the largest disagreement. Thus $L^C$ is a worst-case fit to the archive.

The \emph{full-history oracle} knows $V_C$ and chooses the best feasible state. Its loss and the compression excess are
\begin{align}
F^C(V)&=\min_{y\in C}L^C(V,y),\nonumber\\
\mathcal E^C(V,x)&=L^C(V,x)-F^C(V).
\label{eq:conditional-excess}
\end{align}
Even this oracle may incur loss because the candidates disagree. Subtracting $F^C$ isolates the additional loss caused by compression. The radius $r_C=\sqrt{F^C(V)}$ measures this residual disagreement in image RMS units; it is distinct from observation noise. The oracle state need not be a historical candidate. For $C=X$, write $r_V=r_X$ and omit the superscript on $L,F,\mathcal E$. Compactness of $X,C$ and continuity of the renderers ensure that minima exist.

\begin{table}[t]
\centering\small\setlength{\tabcolsep}{5pt}
\caption{\textbf{A candidate can matter only after correction.} Let $G_1(x)=x$ on $[0,1]$. Both histories initially have the same optimal output. The later observation $C=[0,3/4]$ changes which candidates determine that output.}
\label{tab:running-example}
\begin{tabular}{@{}lll@{}}\toprule
& History 1 & History 2\\\midrule
Candidates & $\{0,1\}$ & $\{0,1/2,1\}$\\
Oracle before correction & $1/2$ & $1/2$\\
Survivors in $C$ & $\{0\}$ & $\{0,1/2\}$\\
Oracle after correction & $0$ & $1/4$\\
\bottomrule\end{tabular}
\end{table}

Table~\ref{tab:running-example} illustrates the information issue. The uncorrected loss is $\max\{x^2,(1-x)^2\}$ for either history; the interior candidate has no effect. After correction, the first history has oracle loss zero, while the second has loss $1/16$. A summary retaining only the original extrema cannot distinguish them. The new observation reveals which region survives, but does not recover the forgotten interior candidate.

Readout sees the whole requested block before emitting frames. Its oracle loss $F^C$ includes the cost of a common state (Supplement Sec.~A.1).

\subsection{What the memory guarantee measures}
An encoder retains at most $B$ history-dependent bits at every prefix, including stored frames, candidates, adaptive weights and observation sets. The public budget $n$ bounds all candidate insertions, including anchors and repetitions. Calibration, current input, $n,H$ and the action clock are public. Query workspace is discarded, and old inputs cannot be replayed. Correction $C$ and the future request arrive after encoding.

For the capacity laws, attributes lie in $X=[0,1]^d$. At most $H$ pre-query actions are clipped expansions
\begin{equation}
A_b(x)=[2x-b]_{[0,1]^d},\qquad b\in\{0,1\}^d,
\label{eq:clipped-actions}
\end{equation}
where brackets clip each coordinate to the unit interval. Their expansion exposes fine distinctions in earlier states. We also permit coordinate assignments: each output coordinate copies an input coordinate, its complement, or a boundary value $0/1$. These include object swaps, reflections and attribute erasure. These assignments may occur arbitrarily often. Every composed renderer pixel is $J$-Lipschitz for fixed $J>0$. For $1<s\le2$, its gradient norm is at most $J$ and, for fixed $M>0$,
\begin{equation}
\|\nabla G_{j,p}(x)-\nabla G_{j,p}(y)\|\le M\|x-y\|^{s-1}.
\label{eq:renderer-regularity}
\end{equation}
Here $s=2$ allows bounded curvature, while $s=1$ requires only the Lipschitz bound. Requested future actions are included in $G_j$; their amplification enters through the uniform constants $J,M$.

We seek the smallest memory guaranteeing $\mathcal E^C\le\delta$ whenever $r_C\le\sigma$. Only the bound $\sigma$ is supplied; $r_C$ is an analysis quantity. The accuracy target $\delta$, radius bound $\sigma$, and promised observation margin $\gamma$ (defined below) are fixed before encoding. A randomized guarantee bounds expected excess for each episode and query fixed independently of its coins. Put
\begin{equation}
q=\delta/\sigma,\qquad \Psi(K,n)=n\log_2(1+K/n).
\label{eq:accuracy-notation}
\end{equation}
Increasing an image radius by $\eta$ adds at most $2\sigma\eta+\eta^2$ to its squared loss, so precision of order $q$ suffices. The function $\Psi(K,n)$ is the order of the bits needed to identify a subset of at most $n$ cells among $K$ cells.

\section{Memory Needed for Later Correction}
\label{sec:refinement}
A correction is \emph{$\gamma$-stable} if moving any archived state by less than $\gamma$ within $X$ cannot change its membership in $C$. A sufficiently small occupied region can then be kept or discarded as a whole. The memory must also describe the candidates within every retained region. Without a positive margin, even a fixed renderer and two candidates can require unbounded memory (Supplement Sec.~E).

\subsection{Memory laws for vector states}
\paragraph{Why coordinate ranges are insufficient.} Consider $V_-=\{(0,0),(1,0),(0,1)\}$ and $V_+=\{(1,1),(1,0),(0,1)\}$. Both have coordinate ranges $[0,1]$, but their hulls are different triangles. For the coupled image channel $G(x)=(x_1+x_2)/2$ and $C=X$, their oracle image values are $1/4$ and $3/4$. A coordinate box loses this distinction. Within a scalar region, two endpoints determine the candidate hull; vector regions need shape information in several directions.

Let $N=2^{dH}\gamma^{-d}$ count the regions that expansion can expose and $W_d=\max\{1,(\gamma/q)^{(d-1)/2}\}$ measure the shape information within one region. Sparse histories encode individual locations; dense histories encode local shapes. Write $B_s^{\rm ref,det}$ and $B_s^{\rm ref,rand}$ for the least deterministic and randomized bit budgets. The symbols $\lesssim,\asymp$ compare rates up to constant factors.

\begin{theorem}[Vector correction memory]
\label{thm:vector-memory}
Fix $d\ge2$ and the regularity bounds. For sufficiently small $\sigma$, $0<\delta\le c\sigma^2$ and $0<\gamma\le c\sigma$, the least deterministic and randomized expected-excess memories satisfy, for $1\le s\le2$ and $2(2^d+1)\le n\le cN$,
\begin{equation}
B_s^{\rm ref,rand}\asymp B_s^{\rm ref,det}
\asymp\Psi\!\left(2^{dH}\max\{\gamma^{-1},q^{-1}\}^{d},n\right).
\label{eq:vector-sparse}
\end{equation}
For $1<s\le2$ and $\gamma\le cq^{1/s}$, a geometric code gives the general upper bound
\begin{equation}
B\lesssim\underbrace{\Psi(N,n)}_{\text{occupied regions}}+\underbrace{\min\{n,N\}W_d}_{\text{local shapes}}.
\label{eq:vector-upper}
\end{equation}
When $n\ge C_0 N W_d$, it matches
\begin{equation}
B_s^{\rm ref,rand}\asymp B_s^{\rm ref,det}\asymp N W_d.
\label{eq:vector-dense}
\end{equation}
The guarantees hold at every prefix and allow any number of coordinate copies, permutations, reflections and boundary resets between at most $H$ clipped expansions. The positive constants $c,C_0$ depend only on $d,s,J,M$.
\end{theorem}

The sparse law allows coupled Lipschitz renderers; coordinates need not act independently. Intermediate archive sizes have the upper bound in Eq.~\eqref{eq:vector-upper}. For $1<s\le2$, $\gamma\le c\min\{\sigma,q^{1/s}\}$ and scalar states with $n\ge6$, two extrema replace local geometry, giving the all-$n$ law
\begin{equation}
B_s^{\rm ref}(d=1)\asymp\Psi(2^H/\gamma,n)
+\min\{n,2^H/\gamma\}\log_2(1+\gamma/q).
\label{eq:scalar-intermediate-law}
\end{equation}
This holds for deterministic and randomized expected excess. For fixed $\sigma$ and static insertion streams, at $\gamma\asymp q^{1/s}$ dense correction requires at least a logarithmic memory factor in one dimension and a polynomial factor $q^{-(d-1)(s-1)/(2s)}$ in higher dimensions, relative to uncorrected rendering. The separation persists under the same uncorrected and corrected disagreement bounds, even when all compared histories have identical old hulls and old loss functions.

For two attributes with bounded curvature ($s=2$), $H=0$ and $\gamma\asymp\sqrt q$, dense correction needs $\Theta_\sigma(q^{-5/4})$ bits. Static uncorrected rendering admits $O_\sigma(q^{-1})$ bits, giving at least a $q^{-1/4}$ correction factor.

\subsection{A geometric code that survives state assignments}
\label{sec:endpoint-code}
CAM measures local shape with fixed exterior probes: each probe records the distance to its nearest historical candidate. Several directions constrain the joint hull. Coordinate copying turns ordinary distance into weighted distance, so the stored bank also includes weighted probes. The probes describe candidate geometry; the renderers $G_j$ retain the general regularity of Sec.~\ref{sec:model}.

Closed cells have side $h_r=h2^{-r}$, with dyadic $h\asymp\gamma$ and $r$ expansions remaining. Each candidate is assigned to one containing cell; $U_i\subseteq[0,1]^d$ gives its normalized coordinates. Actions carry each cell and its boundary assignments into one new cell. Choose $m\asymp\max\{1,\sqrt{\gamma/q}\}$. For integer weights $w\ge0$, $\sum_jw_j\le d$, retain
\begin{align}
Q_{w,z}(u)&=\sum_j(w_ju_j^2-2z_ju_j),\nonumber\\
k_{i,w}(z)&=\left\lfloor m^2\min_{u\in U_i}Q_{w,z}(u)\right\rfloor.
\label{eq:vector-profile}
\end{align}
For full weights, $Q_{\mathbf1,z}(u)=\|u-z\|^2-\|z\|^2$, so the field is a quantized nearest-candidate squared distance with a known offset. Its anchors sample the boundary of $[-K,K+1]^d$ at step $1/m$, with $K=d+2$. Other weights use full grids in at most $d-1$ active coordinates. Copying $u_2\leftarrow u_1$ pulls query weights $(1,1)$ back to $(2,0)$, explaining these auxiliary banks. Supplement Sec.~B specifies their grid bounds and cell indexing.

\paragraph{Exact transport.} In cell coordinates, every permitted action has components $(Au)_j=s_ju_{\pi(j)}+c_j$, with $(s_j,c_j)\in\{(1,0),(-1,1),(0,0),(0,1)\}$. Expanding the quadratic gives a stored query $(w',z')$ and an offset $a$ such that
\begin{align}
Q_{w,z}(Au)&=Q_{w',z'}(u)+a,\nonumber\\
k'_{i',w}(z)&=k_{i,w'}(z')+m^2a.
\label{eq:vector-transport}
\end{align}
The bank is closed: a pullback with all $d$ active coordinates only permutes or reflects full-weight boundary anchors; other pullbacks use the lower-dimensional grids. Since $m^2a$ is an integer, the update commutes with flooring. Insertions and merges take fieldwise minima. Every field remains the exact quantized minimum of the transported candidates, without replay or cumulative rounding.

\paragraph{Finite bits.} The unquantized, offset-adjusted minima in Eq.~\eqref{eq:vector-profile} are concave along each anchor-grid row. For successive integer fields $k_j$ and differences $d_j=k_j-k_{j-1}$,
\begin{equation}
\Delta^2k_j\le2,\quad |d_j|\le2m+1,\quad
\sum_j(2-d_{j+1}+d_j)=O_d(m).
\label{eq:vector-row-code}
\end{equation}
The nonnegative slope deficits have total $O_d(m)$, permitting an $O_d(m)$-bit unary row code including its initial value and difference. For $d\ge2$, the boundary rows and auxiliary banks therefore cost $O_d(m^{d-1})=O_d(W_d)$ bits per cell. Adding the occupied-cell subset code and its headers gives Eq.~\eqref{eq:vector-upper} at every prefix. Supplement Sec.~B counts every field and supplies the decoder. In one dimension, only two exterior probes remain and cost $O(\log(2+m))$ bits.

\paragraph{Image accuracy.} The full-weight profiles decode
\begin{align}
P_i&=\{v\in[0,1]^d:m^2Q_{\mathbf1,z}(v)\ge k_{i,\mathbf1}(z)\ \forall z\},\nonumber\\
U_i&\subseteq P_i\subseteq\operatorname{conv}U_i+m^{-2}B_1.
\label{eq:vector-local-pool}
\end{align}
Here $B_1$ is the closed Euclidean unit ball. Exterior probes bound each decoded point's distance to the candidate hull by $m^{-2}$ (Supplement Sec.~B). The observation margin places every retained cell inside $C$. Taylor expansion then approximates every pool image by a mixture of actual cell images, with common weights across all frames and error
\begin{equation}
\eta_G\le Jh_r/m^2+(M/s)(\sqrt d\,h_r)^s\le q/4.
\label{eq:vector-image-error}
\end{equation}
Minimizing pool loss over $C$ gives $\mathcal E^C\le2r_C\eta_G+\eta_G^2\le\delta$. Only the proof uses mixtures; the generated frames all come from one selected state. Supplement Secs.~B--D give the complete code, geometric bound and converses.

\subsection{Why dimension changes the cost}
The sparse converse places one unknown point in each of several regions. Stable prefix observations read location bits; coordinate queries read finer labels. An augmented-information argument charges these independent bits even with public randomization. A point grid matches this bound, with coordinatewise error unchanged by copies and reflections.

The dense converse encodes independent radial bits in each region. A sphere admits $\Theta_d((\gamma/q)^{(d-1)/2})$ directions separated at angular scale $\sqrt{q/\gamma}$. Moving a point radially by order $q$ changes the support value in just its selected direction. A stable observation isolates that region, and a smooth visual query turns the support change into an oracle-center shift. A wrong bit incurs excess exceeding a fixed multiple of $\delta$. Fano's inequality therefore charges $\Theta(NW_d)$ bits. In a scalar interval there are only two support directions; locating their endpoints instead incurs the logarithm in Eq.~\eqref{eq:scalar-intermediate-law}.

\section{Reading Out a Finite Visual Archive}
\label{sec:decoder}
The same probe fields used for storage also constrain the histories that could have produced them. A field from Eq.~\eqref{eq:vector-profile} says
\begin{equation}
\frac{k}{m^2}\le\min_{u\in U_i}Q_{w,z}(u)<\frac{k+1}{m^2}.
\label{eq:field-evidence}
\end{equation}
Thus every candidate assigned to that cell obeys the lower bound, and at least one obeys the strict upper bound. One number both excludes positions and requires a historical hit in a known region. We call that required hit a \emph{witness}; the field stores no candidate identity.

\emph{Historical excess} measures the extra loss from compression. \emph{Physical risk} measures worst error against possible truths. CAM minimizes physical risk first, using excess only to resolve ties.

\subsection{From stored evidence to historical excess}
Let $Q$ be the candidate pool allowed by a stored message. An occupied-cell record says that the original history contained \emph{at least one} candidate in that cell. More general messages may also identify a mandatory point or a lower bound on the number of candidates. Together these constraints define a family $\mathcal H$ of histories $U\subseteq Q$. For the toy pool $Q=\{0,1/2,1\}$, a message guaranteeing $0,1\in U$ leaves only the interior candidate uncertain (Table~\ref{tab:running-example}). Every history producing a message belongs to its declared family, which can include additional histories.

These constraints are upward closed: adding candidates cannot violate them. Finite conditioning needs no margin: every catalogue point has exact membership in $C$, even when $C$ cuts an occupied cell. Define $Q_{\rm out}=Q\setminus C$. The possible corrected histories are exactly
\begin{equation}
U\in\mathcal H|_C\ \Longleftrightarrow\ U\cup Q_{\rm out}\in\mathcal H,
\qquad\varnothing\ne U\subseteq Q\cap C.
\label{eq:family-refinement-main}
\end{equation}
Any old history producing $U$ is contained in $U\cup Q_{\rm out}$, so upward closure makes this completion feasible; conversely, the completion itself produces $U$. A required region can be witnessed by a deleted point. If it contains any allowed point outside $C$, survivors need not hit it. Otherwise its witness requirement survives. This conditions the declared family exactly; Supplement Sec.~F also handles count constraints.

The worst compression excess allowed by the corrected message is
\begin{equation}
R_C(x)=\max_{U\in\mathcal H|_C}\bigl[L^C(U,x)-F^C(U)\bigr],
\qquad \widehat x\in\arg\min_{x\in C}R_C(x).
\label{eq:finite-risk}
\end{equation}
Both losses use the corrected $U$ and the same output domain $C$. Put $Q_C=Q\cap C$. For a trial output $y$, $a_C(y)$ is the least loss threshold that satisfies all historical constraints. For a candidate $u$, $b_C(u)$ is the least full-history oracle loss compatible with retaining $u$. It measures how much historical disagreement is unavoidable if $u$ remains an explanation:
\begin{align}
a_C(y)&=\min\{t:\{u\in Q_C:\ell_G(u,y)\le t\}\in\mathcal H|_C\},\nonumber\\
b_C(u)&=\min_{y\in C}\max\{\ell_G(u,y),a_C(y)\},\nonumber\\
R_C(x)&=\max_{u\in Q_C}\{\ell_G(u,x)-b_C(u)\}.
\label{eq:computable-readout}
\end{align}
First find the smallest loss sublevel at $y$ that meets all constraints, giving $a_C(y)$. Next force $u$ into that history and optimize over $y$, giving $b_C(u)$. Finally maximize $\ell_G(u,x)-b_C(u)$ over candidates. All witnesses and $u$ must share this same comparator $y$; testing each witness with a different comparator loses their joint constraint.

To verify the middle equality, adjoin $u$ to the first feasible loss sublevel at $y$. Upward closure gives a compatible history with loss at most $\max\{\ell_G(u,y),a_C(y)\}$. Conversely, every compatible history containing $u$ reaches at least that threshold at its oracle. Minimizing over $y$ proves the identity for $b_C$; maximizing over $u$ gives the exact risk. This calculation also accepts a public semantic loss before neural generation.

Every minimization uses the same compact oracle/output domain, which may be $S_{\rm out}\subseteq C$. In the neural test it is the current feasible set $S$, while coarser $C$ filters the archive.

\paragraph{Using the certificate.} Any output with $R_C(x)\le\tau$ ensures $\mathcal E^C(V,x)\le\tau$ for every compatible surviving history. Such an output exists precisely when the minimum of $R_C$ over the chosen output domain is at most $\tau$. The photographic tests enumerate the reachable histories to check this certificate.

\subsection{From historical constraints to possible true states}
The physical decision uses a given bound $f=\sigma^2$ on historical disagreement. Histories with $F^C(U)>f$ are inadmissible. A candidate with $b_C(u)>f$ belongs to none of the remaining histories and can be excluded; those with $b_C(u)\le f$ form the possible-truth set $Q_f$. CAM minimizes the worst image loss over $Q_f$, then historical excess among tied physical optima.

In Table~\ref{tab:running-example}, corrected histories $\{0\}$ and $\{0,1/2\}$ have oracle losses $0$ and $1/16$. Thus $b_C(0)=0$ and $b_C(1/2)=1/16$: only $0$ is possible below this budget, and both are possible at or above it. Formally, let $\mathcal H_f=\{U\in\mathcal H|_C:F^C(U)\le f\}$.

\emph{Joint witnesses.} Let $a,b,c$ be unit-side equilateral image states, with their triangle as $C=X$ and the output domain. Require $a,b$ and leave $c$ optional. With $G(x)=x$, midpoint and circumcenter give normalized two-pixel losses
\begin{align}
F(\{a,b\})&=\tfrac12(\tfrac12)^2=\tfrac18,\nonumber\\
F(\{a,b,c\})&=\tfrac12(\tfrac23\tfrac{\sqrt3}{2})^2=\tfrac16.
\label{eq:joint-witness-floors}
\end{align}
At $f=1/8$, joint constraints exclude $c$; independent pair checks retain it. Their physical risks are $1/8$ and $1/6$ on the same two possible truths (Supplement Sec.~F).

\Needspace{10\baselineskip}
\begin{theorem}[Exact support from historical constraints]
\label{thm:physical-support}
For nonempty $\mathcal H_f$, the already computed oracle profile $b_C$ gives
\begin{align}
Q_f&:=\{u\in Q_C:b_C(u)\le f\}
=\bigcup_{U\in\mathcal H_f}U,\label{eq:truth-support}\\
R_f(x)&:=\max_{U\in\mathcal H_f}[L^C(U,x)-F^C(U)]\nonumber\\
&=\max_{u\in Q_f}[\ell_G(u,x)-b_C(u)],\label{eq:restricted-risk}\\
P_f(x)&:=\max_{U\in\mathcal H_f}\max_{u\in U}\ell_G(u,x)
=\max_{u\in Q_f}\ell_G(u,x).\label{eq:physical-risk}
\end{align}
Thus $P_f$ is the exact worst image loss over the truths allowed by the declared constraints. It bounds true-state loss when $x_{\rm true}\in V_C$ and $F^C(V)\le f$. CAM minimizes $P_f$ over the oracle's output domain, then $R_f$ among tied physical optima.
\end{theorem}

\begin{proof}
If $u\in U\in\mathcal H_f$, then $b_C(u)\le F^C(U)\le f$. Conversely, a history attaining $b_C(u)\le f$ belongs to $\mathcal H_f$ and contains $u$. Restricting to $\mathcal H_f$ therefore leaves the minimum floor of every eligible $u$ unchanged. Interchanging the maxima over histories and their members gives Eq.~\eqref{eq:restricted-risk}; taking their union gives Eq.~\eqref{eq:physical-risk}. Each maximizing state has an attaining compatible history, proving tightness.
\end{proof}

A sound additional current set $S$ restricts the truth support to $Q_f\cap S$. Without an informative bound $f$, the support returns to $Q_C$ and the physical objective becomes outer-pool loss; historical excess still distinguishes tied physical optima. For a covered true state and calibrated images, the useful guarantees are
\begin{align}
\ell_G(x_{\rm true},x)&\le P_f(x)\le f+R_f(x),\nonumber\\
\max_j\|I_j-G_j(x)\|_{P_j}&\le\varepsilon+\sqrt{P_f(x)},
\label{eq:covered-physical}
\end{align}
where $\max_j\|I_j-G_j(x_{\rm true})\|_{P_j}\le\varepsilon$. Calibration error and historical coverage remain distinct from compression. The guarantee covers all requested frames from the chosen common state. Supplement Sec.~F gives the complete threshold sweep and query costs.

\begin{figure*}[t]
\centering\includegraphics[width=\textwidth]{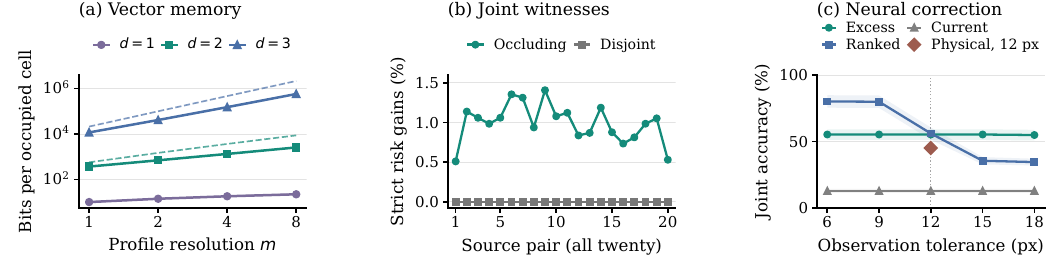}
\caption{\textbf{Memory, physical risk and visual correction.} (a) All vector-code resolutions; dashed curves store fields independently. (b) All critical disagreement bounds: fraction of admissible joint-packet/budget pairs with strictly smaller common worst-truth photo MSE than independent-witness readout. Disjoint images give no improvement. (c) Complete sensor-tolerance sweep for excess readout, ranked retention and current information; physical readout is marked at $12$ pixels. Shading gives pointwise whole-seed $95\%$ intervals.}
\label{fig:diagnostics}
\end{figure*}

\section{Experiments}
\label{sec:experiments}
The three tests evaluate different parts of CAM (Table~\ref{tab:experiment-map}). The geometric diagnostic uses the quadratic-profile code; the image tests use occupancy packets with the finite-constraint readout. All declared COIL-20 object/view cohorts~\cite{nene1996coil} and neural conditions are included.

\begin{table}[t]
\centering\small\setlength{\tabcolsep}{3pt}
\caption{\textbf{Which part of CAM each test evaluates.}}
\label{tab:experiment-map}
\begin{tabular}{@{}p{0.17\columnwidth}p{0.29\columnwidth}p{0.45\columnwidth}@{}}\toprule
Test & Stored packet & What it checks\\\midrule
Geometry & Quadratic fields & Finite bits; exact updates\\
Photos & 4-bit occupancy & Joint witnesses; physical risk\\
Neural & 64-bit occupancy & Hidden-state recovery after actions\\
\bottomrule\end{tabular}
\end{table}

\subsection{Finite bits and exact state evolution}
The vector code is tested in dimensions $1,2,3$ at $m=1,2,4,8$, including its count header, subset rank, padded profile fields and byte carrier. Every transported packet is compared with the literal candidate history. Copies, swaps, reflections, resets and clipped expansions preserve all integer fields exactly. Figure~\ref{fig:diagnostics}a shows the scalar logarithm and vector polynomial growth, with independently stored fields as the coding ablation. Supplement Sec.~H gives the scalar study ($180$ alphabets, $720$ streams, $1{,}440$ stable queries) and photographic endpoint precisions.

\subsection{Physical readout from joint occupancy}
The joint-state test pairs all twenty COIL-20 objects cyclically and uses every starting view. Two three-phase objects overlap on a $128\times192$ canvas, so visibility depends on their joint state. Three frames rotate A, exchange positions while rotating B, then rotate A again from the same initial phase pair. All $511$ histories with an unrestricted filter and six single-object phase filters give $4{,}606{,}560$ nonempty corrected queries; empty cases remain counted. Each phase index is grouped as $\{0,1\}$ or $\{2\}$; four bits record occupied product cells. An equal-bit point code shares all images, observations and outputs. Enumeration verifies both corrected families.

We compare physical readout with point retention, pool minimax, exact excess and independent-witness controls at nine public bounds $f$. A complete sweep of profile breakpoints scores exact and independent outputs on the same truth support. Exact constraints strictly lower worst-truth photo MSE in $2{,}839$ of $285{,}793$ admissible joint-packet/budget contexts, with ties elsewhere and strict gains spanning every source pair (Fig.~\ref{fig:diagnostics}b). The disjoint-photo control has no such improvement. These are structural-spectrum counts. At $f=1$, joint occupancy attains the smallest equal-bit worst-truth bound for both renderers, tying pool and independent readout (Table~\ref{tab:physical-photo}); point retention has lower uniform-truth mean error. Supplement Secs.~G and I give every fixed-budget cohort and the complete single-object study.

\begin{table}[t]
\centering\small\setlength{\tabcolsep}{2.5pt}
\caption{\textbf{Photographic risk at $f=1$.} All $4{,}606{,}560$ nonempty queries per renderer. $P$: worst loss on the common truth support; $e$: uniform-truth mean within each history. Losses are scaled by $10^3$ (lower is better). Bold compares four-bit codes; the nine-bit reference is separate.}
\label{tab:physical-photo}
\begin{tabular}{@{}lrrrrr@{}}\toprule
& & \multicolumn{2}{c}{Occluding} & \multicolumn{2}{c}{Disjoint}\\
Method & Bits & $P\downarrow$ & $e\downarrow$ & $P\downarrow$ & $e\downarrow$\\\midrule
Joint: pool & 4 & \textbf{56.313} & 35.599 & \textbf{51.208} & 31.436\\
Joint: excess & 4 & 56.398 & 35.513 & 51.210 & 31.431\\
Joint: physical & 4 & \textbf{56.313} & 35.610 & \textbf{51.208} & 31.436\\
Joint: independent & 4 & \textbf{56.313} & 35.610 & \textbf{51.208} & 31.436\\
Point: native & 4 & 60.060 & \textbf{31.745} & 53.620 & \textbf{28.160}\\
Point: pool & 4 & 58.214 & 35.767 & 53.026 & 30.299\\
Point: excess & 4 & 59.761 & 31.904 & 53.418 & 28.384\\
Point: physical & 4 & 58.214 & 35.515 & 53.026 & 28.965\\
Point: independent & 4 & 58.214 & 35.515 & 53.026 & 28.965\\
\midrule
Full-history oracle & 9 & \textbf{38.233} & \textbf{24.164} & \textbf{34.614} & \textbf{19.328}\\
\bottomrule\end{tabular}
\end{table}

\begin{figure*}[t]
\centering\includegraphics[width=\textwidth]{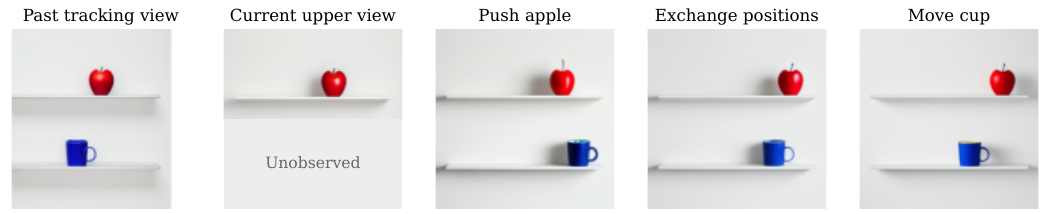}
\caption{\textbf{Coherent continuation with physical readout.} A tracked past view and current upper view constrain the joint state. All three actual neural outputs are uncropped; grey marks unobserved input. The requested actions move one object, exchange positions, then move the other.}
\label{fig:neural-case}
\end{figure*}

\subsection{State correction in neural continuations}
Frozen Stable Diffusion with Canny ControlNet and LCM-LoRA~\cite{rombach2022latent,zhang2023controlnet,luo2023lcmlora} renders an apple and cup from joint layouts and edges; frozen RT-DETR-R18~\cite{zhao2024rtdetr} reads candidates from input images. Three tracking views interleave known moves and exchange, providing relative positions for four later absolute placements. We cross $16$ layouts, $32$ seeds and four conditions: full/upper current views and preserving/erasing requests. Each request produces three frames; erasing aligns the objects before output. The $16\times16$ joint-position grid uses $2\times2$ occupancy cells, giving an $8\times8$, $64$-bit archive decoded by Sec.~\ref{sec:decoder}.

Every readout outputs in the raw current set $S$; its coarse-cell closure $C$ filters history. We use $f=1$, so physical readout minimizes loss against the truth pool $S\cap Q_C$, then resolves ties by historical excess. Sound-pool minimax uses the same truth pool. Generator, parser, noise and sampling are shared. The zero-history prior uses public $P_{\rm layout}\cap S$, falling back to $S$ if empty.

\paragraph{Images and failures.} Position loss is worst-frame squared object RMS against the prescribed trajectory. Missing, duplicated, or invalid targets incur unit per-object error; no episode is removed. A colour-only parser supplies a separate location check. For selected-state loss $e_{\rm state}$, parsed-image loss $e_{\rm img}$, and parsed-to-command RMS error $\rho$ under this same failure metric,
\begin{equation}
\left|\sqrt{e_{\rm img}}-\sqrt{e_{\rm state}}\right|\le\rho.
\label{eq:neural-transfer-main}
\end{equation}
The triangle inequality proves this relation; $\rho$ is measured after generation. Joint accuracy requires both objects within one grid step in every frame.

In each condition, current sets cover $512/512$ prescribed states and transported surviving histories cover $504/512$; eight empty histories use current-set fallback. Under an upper view, memory reduces mean candidate count from $18.41$ to $4.74$. For preserving requests, physical readout lowers true-state selection loss against current minimax by $0.01950$, with whole-seed $95\%$ interval $[0.01886,0.02005]$. This paired comparison precedes rendering and separates state recovery from the generator's control error. Erasing the hidden coordinate largely removes this gain: the same history is then queried by frames that no longer depend on the hidden relation. Independent development calibration envelopes cover $97.12\%/99.15\%$ of test RT-DETR/HSV residuals, including failures (Supplement Sec.~J).

For occluded preserving requests, joint conditioning improves state recovery over ranked, factorized and current-only memory. Outer-pool and sound-pool remain close to CAM; physical and sound-pool readouts differ on only seven queries. Ranked retention has lower detector loss and higher threshold accuracy; current-only outputs have lower detector loss and fewer invalid detections. The photographic tests isolate the extra value of historical witnesses. Figure~\ref{fig:diagnostics}c and Supplement Sec.~J give all fifteen methods, state and image losses, both parsers and accuracy across every condition.

\paragraph{Computation.} Let $N_c,N_o$ count candidates and outputs, and $\Lambda$ count cap memberships and mandatory points. Forming pixel losses costs $O(TN_oN_cP)$ plus rendering; the oracle profile costs $O(N_o(N_c+\Lambda))$. Two-pass readout with $b$-output blocks uses $O(bN_c+N_c+N_o+\Lambda)$ workspace. The $64$-bit packet is separate from its $512$\,KiB public loss table. Profile-inclusive excess/physical queries take $0.95/0.98$\,ms, excluding table construction; shared generation takes $4.22$\,s/frame with six steps (Supplement Secs.~F and J).

\paragraph{Case study.} Figure~\ref{fig:neural-case} tracks the hidden cup through exchange. The current view shows the apple but leaves the cup unobserved. Past tracking constrains their horizontal offset, letting the visible apple narrow the cup's possible positions before future actions are applied. Current-only minimax misplaces it; after exchange, its cup error becomes an apple error. CAM preserves and transports their joint relation. One initial state then explains all three frames.

\section{Conclusion}
We studied finite memory for correctable visual continuation. Matching sparse and dense vector laws reveal a dimension-dependent correction cost. Our geometric summaries update exactly under the stated actions. Historical witnesses identify possible truths and support physical-risk minimization under bounded disagreement. Photographic and neural tests validate joint decisions and recovery of occluded positions.

\clearpage
\section*{LLM Usage Statement}
LLM was used for language polishing. All AI-assisted parts were reviewed and verified by the authors, who take full responsibility for the content of this paper.

% Main bibliography: original entries, inlined.

% Start the supplementary document on a fresh page with its original numbering.
\clearpage
\renewcommand*{\HyperDestNameFilter}[1]{supplement.#1}
\setcounter{page}{1}
\setcounter{figure}{0}
\setcounter{table}{0}
\setcounter{footnote}{0}
\setcounter{dbltopnumber}{3}
\renewcommand{\dbltopfraction}{0.95}
\BeforeBeginEnvironment{corollary}{\Needspace{5\baselineskip}}
\BeforeBeginEnvironment{supptheorem}{\Needspace{5\baselineskip}}
\BeforeBeginEnvironment{supplemma}{\Needspace{5\baselineskip}}
\BeforeBeginEnvironment{suppprop}{\Needspace{5\baselineskip}}
\BeforeBeginEnvironment{suppcorollary}{\Needspace{5\baselineskip}}
% Supplementary material: original text with internal keys namespaced.

\lefthyphenmin=2 \righthyphenmin=3 \maketitlesupplementary \appendix \numberwithin{equation}{section} \newtheorem{supplemma}{Lemma}[section] \newtheorem{suppprop}[supplemma]{Proposition} \newtheorem{suppcorollary}[supplemma]{Corollary} \newtheorem{supptheorem}[supplemma]{Theorem}

Section A separates the cost of a common state from compression error. Section B derives the vector representation, exact updates, bit count and reconstruction bound; Section C proves the sparse and dense lower bounds. Section D supplies the uncorrected reference, and Section E establishes the scalar rate and the need for observation clearance. Section F derives conditional readout, physical-risk guarantees and computational costs. Section G specifies the two-object photographic tests, while Section H details encoding precision and causal transport. Section I analyzes single-object photographs. Section J gives the neural protocol, error metric, calibration, coverage and sensitivity results.

\section{Visual Evidence, Tasks, and Memory}
\label{app:model}
We use the main definitions of $L^C,F^C,\mathcal E^C$ and the memory budgets. The symbol $B_1=\{v:\|v\|_2\le1\}$ denotes the closed Euclidean unit ball. Positive constants $c,C$ may change between inequalities; their dependence is specified with each bound.

\subsection{Common-state and compression costs}
\label{app:common-state-cost}
On a common nonempty compact output domain $S_{\rm out}$, put $F_{\rm coh}=\min_{x\in S_{\rm out}}L^C(V,x)$ and
\begin{equation}
F_{\rm ind}=\max_j\min_{y\in S_{\rm out}}\max_{u\in V_C}
\|G_j(u)-G_j(y)\|_{P_j}^2.
\end{equation}
The max--min inequality gives $F_{\rm ind}\le F_{\rm coh}$, hence
\begin{equation}
L^C(V,x)-F_{\rm ind}
=(F_{\rm coh}-F_{\rm ind})+[L^C(V,x)-F_{\rm coh}],
\end{equation}
with both terms nonnegative. The second bracket is compression excess on the chosen output domain. The first is the oracle cost of requiring one shared state across all frames.

\section{Finite Vector Archives}
\label{app:archive-geometry}
\subsection{A finite vector archive and its exact updates}
\label{app:vector-upper}
We prove the upper bound with $q=\delta/\sigma$. All constants may depend on fixed $d,s,J,M$. Take the largest admissible reciprocal dyadic width $h$ and the smallest integer $m\ge1$ satisfying
\begin{equation}
\begin{gathered}
h\le\frac{\gamma}{4\sqrt d},\qquad
\frac{M}{s}d^{s/2}h^s\le\frac q8,\\
m^2\ge\frac{8Jh}{q}.
\end{gathered}
\label{eq:vector-resolution}
\end{equation}
In the stated margin regime, $h\asymp\gamma$ and $m\asymp\sqrt{\max\{1,\gamma/q\}}$. With $r$ expansions remaining, let $L_r=2^r/h$ and $h_r=L_r^{-1}$. These values and the action clock are public.

\paragraph{The retained sets form a cover.} Use the closed cells $C_i=(i+[0,1]^d)/L_r$, $i\in\{0,\ldots,L_r-1\}^d$. Each insertion is initially assigned to one containing cell. For every occupied cell, an implicit finite set $U_i\subseteq[0,1]^d$ describes its normalized candidates, with
\begin{equation}
V=\bigcup_{i\ {\rm occupied}}\frac{i+U_i}{L_r}.
\label{eq:vector-cover}
\end{equation}
The sets $U_i$ track assigned candidates in the proof; the persistent state consists of their integer profiles. A boundary candidate may remain assigned to either containing cell. Each action maps an old cell into one destination, transforms its assigned set and merges incoming sets by union. Thus at most $n$ cells are occupied.

Put $K=d+2$ and define the public weight family $\mathcal W=\{w\in\mathbb Z_{\ge0}^d:\sum_jw_j\le d\}$. For $w\ne0$, retain the integer profile
\begin{align}
Q_{w,z}(u)&=\sum_j(w_ju_j^2-2z_ju_j),\nonumber\\
k_{i,w}(z)&=\left\lfloor m^2\min_{u\in U_i}Q_{w,z}(u)\right\rfloor.
\label{supp-eq:vector-profile}
\end{align}
If $w=\mathbf1$, anchors $z$ range over the boundary lattice of $[-K,K+1]^d$ with step $1/m$. For every other nonzero $w$, use the full lattice with $z_j\in[-Kw_j,(K+1)w_j]\cap m^{-1}\mathbb Z$; inactive coordinates have $z_j=0$. The zero-weight profile is identically zero. Only $w=\mathbf1$ can have $d$ active coordinates, since the integer weights sum to at most $d$. Insertion takes the minimum of each old field and its directly quantized value at the new point.

\paragraph{Action closure.} In local coordinates, an allowed coordinate assignment has components $u'_j=s_ju_{\pi(j)}+c_j$, with $(s_j,c_j)=(1,0),(-1,1),(0,0)$ or $(0,1)$. Repeated sources implement copying. Copies and reflections send cell index $i_j$ to $i_{\pi(j)}$ and $L_r-1-i_{\pi(j)}$, respectively; resets select the first or last cell. A clipped expansion decreases $r$ and sends an old index to
\begin{equation}
i'_j=[i_j-b_jL_r/2]_{\{0,\ldots,L_r/2-1\}}.
\end{equation}
Its local coordinate is unchanged when $(b_j=0,i_j<L_r/2)$ or $(b_j=1,i_j\ge L_r/2)$; otherwise it is the constant $1-b_j$. Thus it has the same local algebra. Closed cells make these statements valid at their boundaries.

For a target field $(w,z)$, direct expansion gives
\begin{align}
w'_k&=\sum_{j:\pi(j)=k,\ s_j\ne0}w_j,\nonumber\\
z'_k&=\sum_{j:\pi(j)=k,\ s_j\ne0}s_j(z_j-w_jc_j),\nonumber\\
a&=\sum_j(w_jc_j^2-2z_jc_j),\nonumber\\
Q_{w,z}(u')&=Q_{w',z'}(u)+a.
\label{eq:vector-assignment-algebra}
\end{align}
Indeed, each nonconstant term expands as $w_j(s_ju+c_j)^2-2z_j(s_ju+c_j)=w_ju^2-2s_j(z_j-w_jc_j)u+w_jc_j^2-2z_jc_j$; reset terms are constant. We have $w'\in\mathcal W$ and $z'\in m^{-1}\mathbb Z^d$. A reflected ratio $z_j/w_j$ becomes $1-z_j/w_j$, and merging copied coordinates takes a positive-weighted average. Both operations preserve $[-K,K+1]$. If all coordinates of $w'$ are positive, then $w'=\mathbf1$, the old weight is also $\mathbf1$, and the source map is a bijection without resets. Its anchor is only permuted or reflected, preserving the boundary of $[-K,K+1]^d$. Otherwise its full lower-dimensional bank is present. Finally $m^2a\in\mathbb Z$, so
\begin{equation}
k'_{i',w}(z)=k_{i,w'}(z')+m^2a.
\label{eq:vector-exact-update}
\end{equation}
Incoming entries merge by pointwise minima. Floor commutes with these minima and with the integer offset, preserving every raw-set profile exactly. Arbitrarily many assignments require neither rerounding nor a stored action word.

\subsection{Literal finite-bit accounting}
\label{app:vector-bits}
For fixed $w$, the unquantized minimum in Eq.~\eqref{supp-eq:vector-profile} is concave in $z$. Along an active-coordinate lattice row, let its values be $F_j$ and its integer fields be $k_j$. Since $u_j\in[0,1]$,
\begin{equation}
\begin{aligned}
\Delta^2F_j&\le0,& |F_j-F_{j-1}|&\le2/m,\\
\Delta^2k_j&\le2,& |k_j-k_{j-1}|&\le S:=2m+1.
\end{aligned}
\label{eq:vector-row-bounds}
\end{equation}
Also $|k_j|\le R_m:=d(2K+3)m^2$, including negative fields. Store $k_0+R_m$ in $\lceil\log_2(2R_m+1)\rceil$ bits. For a row of length $\ell\ge2$, store $d_1+S$, where $d_j=k_j-k_{j-1}$, in $\lceil\log_2(2S+1)\rceil$ bits. The remaining nonnegative deficits $t_j=2-d_{j+1}+d_j$, $1\le j\le\ell-2$, obey
\begin{equation}
\sum_jt_j=2(\ell-2)+d_1-d_{\ell-1}\le2(\ell-2)+2S.
\end{equation}
Encode $t_j$ as $t_j$ ones and a zero. Their total length is at most $3(\ell-2)+2S$. Pad each row to this public bound. Its fields reconstruct all slopes and values exactly. A one-entry row stores only its first value.

For $d\ge2$, the full-weight boundary has $O_d(m^{d-2})$ rows of length $O_d(m)$, including harmless repeated face-edge anchors. A proper weight with $k\le d-1$ active coordinates has $O_d(m^{k-1})$ full-grid rows. There are $\binom{2d}{d}$ public weight vectors. Thus the complete padded cell field has
\begin{equation}
P_m=O_d(m^{d-1})\quad(d\ge2),\qquad
P_m=O(\log(2+m))\quad(d=1).
\label{eq:vector-profile-bits}
\end{equation}
The scalar case has only two exterior anchors. The unary count is what removes a precision logarithm per anchor.

Let $N_r=L_r^d$ and $k_r=\min(n,N_r)$. A literal packet contains one leading framing bit, a $\lceil\log_2(k_r+1)\rceil$-bit occupied-cell count $k$, the colexicographic rank of its subset in $\lceil\log_2\binom{N_r}{k}\rceil$ bits, and its $k$ fixed-length cell fields. The framing bit locates the header without an uncharged length field. The maximum native length is
\begin{equation}
\begin{aligned}
B_r={}&1+\lceil\log_2(k_r+1)\rceil\\
&+\max_{0\le k\le k_r}\left\{\left\lceil\log_2\binom{N_r}{k}\right\rceil+kP_m\right\}.
\end{aligned}
\label{eq:vector-literal-bits}
\end{equation}
The subset-count bound gives $B_r=O(\Psi(N_r,n)+\min(n,N_r)P_m)$ for $n\ge1$. Decoding, updating and re-encoding use temporary workspace. Only this integer packet persists. Since $N_r\le N_H\asymp2^{dH}\gamma^{-d}$, Eq.~\eqref{eq:vector-profile-bits} proves the asserted worst-prefix bit bound once its error is established.

\subsection{A containing pool and the visual error bound}
\label{app:vector-image-bound}
For a full-weight anchor $a$, $m^2\|a\|^2$ is integral, so $\widetilde D_i(a)=m^{-2}k_{i,\mathbf1}(a)+\|a\|^2$ is exactly $m^{-2}\lfloor m^2\min_{u\in U_i}\|a-u\|^2\rfloor$. Decode the compact pool over these boundary anchors:
\begin{equation}
P_i=\{x\in[0,1]^d:\|a-x\|^2\ge\widetilde D_i(a)\ \forall a\}.
\end{equation}
It contains $U_i$. To bound its reverse error, take $x\in P_i\setminus\operatorname{conv}U_i$, let $y$ be its hull projection, $t=\|x-y\|$ and $v=(x-y)/t$. Projection gives $v^\top(x-u)\ge t$ for every $u\in U_i$. The ray $x+\alpha v$ hits the exterior cube at $\alpha\ge K$. A nearest same-face lattice anchor is $a'=x+\alpha v+e$, with $\|e\|\le\eta_0=\sqrt{d-1}/(2m)$. Therefore
\begin{align}
\min_{u\in U_i}\|a'-u\|^2&\ge(\alpha+v^\top e+t)^2,\nonumber\\
\|a'-x\|^2&\ge\min_{u\in U_i}\|a'-u\|^2-m^{-2}.
\end{align}
The first inequality uses $\alpha+v^\top e>0$. Expanding the second and canceling $(\alpha+v^\top e)^2$ gives
\begin{align}
2(\alpha+v^\top e)t+t^2&\le\eta_0^2+m^{-2},\nonumber\\
t&\le\frac{(d-1)/4+1}{2(K-\sqrt{d-1}/2)}m^{-2}\le m^{-2}.
\label{eq:vector-local-hull-error}
\end{align}
The case $x\in\operatorname{conv}U_i$ is immediate. The last inequality holds for $K=d+2$.

Rescaling a decoded point to its physical cell yields a mixture $Y$ of actual assigned states with $\|\mathbb EY-x\|\le h_r/m^2$ and $\|Y-x\|\le\sqrt d h_r$. Integral Taylor expansion of any composed renderer pixel $f$ gives
\begin{align}
|f(x)-\mathbb Ef(Y)|
&\le J\|x-\mathbb EY\|+(M/s)\mathbb E\|Y-x\|^s\nonumber\\
&\le Jh_r/m^2+(M/s)d^{s/2}h_r^s\le q/4=:e.
\label{eq:vector-shared-mixture}
\end{align}
The Taylor remainder follows by integrating $[\nabla f(x+t(Y-x))-\nabla f(x)]^\top(Y-x)$ from zero to one. The mixture weights do not depend on the pixel or requested frame.

Test the correction at each occupied cell's center. Its distance to every assigned state is at most $\sqrt d h_r/2<\gamma$, so the result is their common membership. Every decoded point in a retained cell is within $\sqrt d h_r<\gamma$ of an actual assigned state. The physical pool $P_C=\bigcup_{i\ {\rm retained}}(i+P_i)/L_r$ consequently satisfies $V_C\subseteq P_C\subseteq C$. For the full-history oracle $x_*$, every mixture member lies within image radius $r_C$ of its sequence. Convexity of the product norm $\max_j\|\cdot\|_{P_j}$ and Eq.~\eqref{eq:vector-shared-mixture} imply $L(P_C,x_*)\le(r_C+e)^2$. Minimizing pool loss over the same compact output domain as the oracle gives
\begin{align}
L(V_C,\widehat x)&\le L(P_C,\widehat x)\le L(P_C,x_*)\le(r_C+e)^2,\nonumber\\
\mathcal E^C(V,\widehat x)&\le2r_Ce+e^2
\le\delta/2+\delta^2/(16\sigma^2)\le\delta.
\end{align}
All output frames still come from the single state $\widehat x$. This proves the vector upper bound. Alternatively, the ranked proxy grid at coordinate resolution $2^{-r}\min\{\gamma,q\}$ gives $O(\Psi(2^{dH}\max\{\gamma^{-1},q^{-1}\}^d,n))$ bits by the pairing invariant proved in Sec.~\ref{app:vector-lower}. Coordinate assignments do not increase its maximum coordinate error, so this alternative also permits arbitrarily many assignments. Taking the smaller code gives the stated general upper bound.

\subsection{From the same fields to finite-history readout}
\label{app:vector-catalogue-caps}
Let a public finite catalogue $\mathcal X$ contain the actual historical states. For each occupied cell, keep the catalogue points in its closed cell satisfying
\begin{equation}
\begin{aligned}
Q_i=\bigl\{x\in\mathcal X\cap C_i:{}&\\
\left\lfloor m^2Q_{w,z}(L_rx-i)\right\rfloor&\ge k_{i,w}(z)\ \forall(w,z)\bigr\}.
\end{aligned}
\end{equation}
Every actual assigned state lies in $Q_i$. Moreover each retained minimum supplies the required cap
\begin{equation}
W_{i,w,z}=\left\{x\in Q_i:
\left\lfloor m^2Q_{w,z}(L_rx-i)\right\rfloor=k_{i,w}(z)\right\}.
\end{equation}
Some assigned state attains that quantized minimum, so the actual history hits every cap and every nonempty cell pool. Their union $Q=\bigcup_iQ_i$ and these hitting constraints define an upward-closed family containing the actual history. The caps are computed from the same fields and need no additional witness bits. Cell ownership and the insertion budget may make the reachable encoder fiber smaller. The readout identities in Sec.~\ref{app:finite-loss} are exact for this declared family, which contains the actual history.

\subsection{A reusable static hull code}
\label{app:static-hull-reuse}
\begin{supplemma}[Finite recurrent hull approximation]
\label{lem:static-hull-profile}
For fixed $p\ge2$, $R>0$ and $0<\epsilon\le R$, a nonempty finite archive $U\subset B_R^p$ admits an insertion-updatable code of $O_p((R/\epsilon)^{(p-1)/2})$ bits whose decoded convex body has Hausdorff distance at most $\epsilon$ from $\operatorname{conv}U$.
\end{supplemma}
\begin{proof}
Normalize $u\mapsto(u+R\mathbf1)/(2R)\in[0,1]^p$. Use only the full-weight boundary bank above, with $K=p+2$ and $m=\lceil\sqrt{2R/\epsilon}\rceil$. Its fields update by minima on insertion. The geometric argument of Eq.~\eqref{eq:vector-local-hull-error} gives $U'\subseteq P\subseteq\operatorname{conv}U'+m^{-2}B_1$. Taking convex hulls preserves both inclusions, so rescaling gives Hausdorff error at most $2R/m^2\le\epsilon$. The boundary-row argument gives $O_p(m^{p-1})$ literal bits, with one additional empty-archive flag. Thus the bound holds at every insertion prefix. Applying this lemma in lifted dimension $p=d+1$ supplies the static geometric code used by the uncorrected comparison.
\end{proof}
\section{Information Required for Correction}
\subsection{Randomized converses in arbitrary dimension}
\label{app:vector-lower}
Put $A=2^{dH}$, $N_\gamma=A\gamma^{-d}$ and $q=\delta/\sigma$. All constants below may depend on fixed $d,s,J,M$. We take $\sigma$ sufficiently small, $\delta\le c\sigma^2$ and $\gamma\le c\sigma$. Each construction has a common insertion count and ends with the origin, so neither stream length nor the current input reveals a hidden label. All random coins, including later readout coins, are placed on an independent public tape $\mathsf R$.

We first record the information step. If independent fair bits $W_1,\ldots,W_b$ can each be decoded from the packet $S$, $\mathsf R$ and earlier bits, with error at most $p<1/2$, the entropy chain rule gives
\begin{align}
B&\ge H(S\mid\mathsf R)\ge I(W;S\mid\mathsf R)\nonumber\\
&=b-\sum_{i=1}^bH(W_i\mid S,\mathsf R,W_{<i})
\ge b[1-h_2(p)].
\label{eq:vector-augmented-information}
\end{align}
Indeed the binary decoding-error indicator determines $W_i$ from its estimate; its entropy is at most $h_2(p)$. Supplied earlier bits may determine actions and filters, but these choices remain independent of $\mathsf R$.

\paragraph{Sparse histories.}
Assume $2(2^d+1)\le n\le cN_\gamma$. We prove
\begin{equation}
B^{\rm ref,rand}_s\gtrsim
\Psi\!\left(A\max\{q^{-1},\gamma^{-1}\}^{d},n\right),
\qquad 1\le s\le2.
\label{eq:vector-sparse-lower}
\end{equation}
For occupancy, use $D_0(x)=1/2+\sigma\psi_0(x_1)$, where $\psi_0(t)=(1-8t)_+^3$, and put $C_{\rm b}=[0,1/32]^d$. The origin has intensity $1/2+\sigma$, while all selected nonzero candidates have first coordinate at least $1/8$ and intensity $1/2$. Their midpoint is attained in $C_{\rm b}$ because $\psi_0(0)=1$ and $\psi_0(1/32)=27/64<1/2$. A filter leaving $\{0\}$ or $\{0,u\}$ therefore reads one bit: classification at $1/2+3\sigma/4$ incurs excess at least $\sigma^2/16$ on error. Expected excess $\delta\le\sigma^2/128$ bounds its error by $1/8$. This fixed $C^2$ request satisfies the prescribed regularity for small $\sigma$.

Let $a=n-(2^d+1)\ge n/2$; the other insertions are the $2^d$ cube corners and a final origin. Choose $L_\gamma=2^k=\Theta(\gamma^{-1})$ ordered centers in $[7/16,9/16]$, separated by at least $8\gamma$, with $k\ge5$.

If $a\le A$, set $r=\lceil(\log_2a)/d\rceil\le H$ and choose $a$ public depth-$r$ dyadic cells. In every coordinate of each cell, encode $\ell=\lfloor(H-r)/3\rfloor$ independent coarse bits by $0\mapsto001$, $1\mapsto110$, followed by an independent $k$-bit center index. With public cell index $\kappa$, concatenated coarse-word integers $v_j$, and center indices $z_j$, publish
\begin{equation}
p_j=2^{-r}\bigl[\kappa_j+2^{-3\ell}(v_j+c_{z_j})\bigr],
\qquad j=1,\ldots,d.
\label{eq:vector-sparse-word}
\end{equation}
Order coarse bits by rounds and then coordinates. To read coordinate $j$ in round $t$, earlier bits supply every coordinate's first $t-1$ rounds. Apply the public $r$-step cell selector and those $3(t-1)$ prefix actions. The selected point's coordinates lie in $[1/8,1/4]$ or $[3/4,7/8]$. Query $C_{\rm b}\cup B_j$, where
\begin{equation}
B_j=\{x:x_j\in[1/16,5/16],\ x_i\in[1/16,15/16]\ (i\ne j)\}.
\end{equation}
The point survives exactly for bit zero. Other public cells have a face coordinate and cannot enter $B_j$. After these selectors every coordinate is $0$, $1$ or a current code tail; consequently a nonselected point can enter $C_{\rm b}$ only as the origin. All memberships have clearance at least $1/32$, exceeding $\gamma$.

After all coarse rounds, a selected point has central coordinates $c_{z_j}$. Read each center-index bit with the lower half of its supplied prefix interval, using boundaries halfway between centers and extending an outer boundary by $4\gamma$. Other coordinates range over $[3/8,5/8]$. Every membership has clearance at least $4\gamma$, and the total action count is $r+3\ell\le H$. Neither this filter nor the earlier coarse filter discloses the queried bit. Equation~\eqref{eq:vector-augmented-information} yields
\begin{equation}
B\ge[1-h_2(1/8)]ad(\ell+k)
\gtrsim n\log_2(1+N_\gamma/n).
\label{eq:vector-sparse-first-branch}
\end{equation}
For the last inequality, $dr\in[\log_2a,\log_2a+d)$ and $H-r\le3\ell+2$; the fixed additive terms are absorbed by $k\ge5$.

If $a>A$, put $b=\lfloor a/A\rfloor$ and $g=2^{\lfloor(\log_2b)/d\rfloor}$, so $b/2^d<g^d\le b$. In each depth-$H$ cell divide the central centers into $g$ public consecutive groups per coordinate. Put one point in each Cartesian group, with $Q=\Theta(\gamma^{-1}/g)$ power-of-two alternatives independently in every coordinate. Taking the constant in $n\le cN_\gamma$ small ensures $Q\ge4$. These $Ag^d\le a$ points use the same number of insertions in every history; pad with origins. An $H$-step selector and a rectangular correction read a coordinate-prefix bit inside its public group. Other groups are excluded by at least $4\gamma$, and other dyadic cells again land on faces. Thus
\begin{equation}
B\gtrsim Ag^d d\log_2Q
\gtrsim n\log_2(1+N_\gamma/n).
\label{eq:vector-sparse-second-branch}
\end{equation}
Here $Ag^d=\Theta_d(n)$ and $d\log Q=\Theta_d(\log(N_\gamma/n))$. This proves the randomized occupancy term in both branches.

For precision, publicly occupy $a$ slots among $\Theta_d(N_\gamma)$ separated slots. In each, choose independent coordinate labels $z_j\in\{0,\Delta,\ldots,(L-1)\Delta\}$, where $\Delta=1024q/\lambda_0$ and $L=1+\lfloor\gamma/\Delta\rfloor$. The small constant $\lambda_0>0$ is fixed before encoding. A dyadic selector and the filter
\begin{equation}
[0,5/16]^d\ \cup\ \prod_i[c_i-3\gamma,c_i+3\gamma]
\end{equation}
leave exactly $\{0,c+z\}$; centers are $12\gamma$ apart, so selected and excluded points have clearance at least $2\gamma$ and $8\gamma$. Query
\begin{equation}
D_{c,j}(x)=\tfrac12+\lambda_0x_j(x_j-c_j)+\sigma\psi_0(x_j).
\end{equation}
Its derivative bounds are uniform in $c$ and $j$. For small $\gamma/\sigma$, the low intensity lies in $[1/2,1/2+\sigma/8]$, the radius lies in $[7\sigma/16,\sigma/2]$, and the oracle midpoint is attained in the origin component. Adjacent labels move that midpoint by at least $7\lambda_0\Delta/32$. Nearest-midpoint misclassification therefore incurs at least
\begin{equation}
(\sigma/2)(7\lambda_0\Delta/64)=56\delta.
\end{equation}
Expected excess implies label error at most $1/56$. Conditional Fano and the entropy chain rule give, when $L\ge2$,
\begin{align}
B&\ge ad\bigl[\log_2L-h_2(1/56)
-(1/56)\log_2(L-1)\bigr]\nonumber\\
&\gtrsim n\log_2L.
\label{eq:vector-sparse-fano}
\end{align}
When $\gamma/q$ exceeds a fixed constant this is $\Omega(n\log(1+\gamma/q))$; otherwise that term is absorbed by occupancy. Finally, for $n\le cN_\gamma$,
\begin{equation}
\Psi(N_\gamma\max\{1,\gamma/q\}^{d},n)
\asymp\Psi(N_\gamma,n)+n\log_2(1+\gamma/q).
\end{equation}
Taking the maximum of the two lower bounds proves Eq.~\eqref{eq:vector-sparse-lower}. For the matching ranked-grid upper, use denominator $m_{\rm g}2^r$, choosing $m_{\rm g}\asymp_d\max\{\gamma^{-1},q^{-1}\}$ sufficiently large that $\xi=\sqrt d/(2m_{\rm g})\le\min\{\gamma/2,q/(16J)\}$. An insertion's coordinate error is at most $1/(2m_{\rm g}2^r)$. Assignments preserve this $\ell_\infty$ bound; each expansion doubles it while decreasing $r$. Thus the Euclidean pairing error never exceeds $\xi$, stable filtering preserves the pairing, and the radius comparison gives
\begin{equation}
\mathcal E^C\le4r_CJ\xi+4J^2\xi^2\le\delta.
\end{equation}
The occupied proxy set has the stated ranked-subset cost. Neither error nor memory grows with the number of coordinate assignments.

\paragraph{Dense histories.}
For $d\ge2$ and $\Phi_d(t)=(1+t)^{(d-1)/2}$, we prove
\begin{equation}
B^{\rm ref,rand}_s\gtrsim N_\gamma\Phi_d(\gamma/q)
\quad\text{if}\quad n\ge C N_\gamma\Phi_d(\gamma/q).
\label{eq:vector-dense-lower}
\end{equation}
This uses the common upper-bound regime $\gamma\le c\min\{\sigma,q^{1/s}\}$, $1<s\le2$. If $\gamma/q$ is bounded, the occupancy construction with $\Theta(N_\gamma)$ points and origin padding proves the claim. Otherwise put $\rho=\gamma/4$ and $a_0=128q/\lambda_0$. The ratio is large enough that $a_0\le\rho/2$. Choose a packing of unit vectors in a fixed spherical patch of diameter at most one, with
\begin{equation}
K_0\asymp_d(\rho/a_0)^{(d-1)/2},\qquad
\|v_j-v_k\|\ge2\sqrt{a_0/\rho}\quad(j\ne k).
\end{equation}
Such a packing follows by placing a lattice in a fixed $(d-1)$-dimensional coordinate chart of the sphere. Its patch ensures $v_j\cdot v_k\ge1/2$.

In each depth-$H$ dyadic cell, use central centers $c$ separated by $12\gamma$. Their total number is $N_0=\Theta_d(N_\gamma)$. At each slot independently encode $K_0$ fair bits by the candidates
\begin{equation}
p_{k,c,j}=2^{-H}\bigl[k+c+(\rho-a_0(1-W_{k,c,j}))v_j\bigr].
\label{eq:vector-cap-insertions}
\end{equation}
Add all cube corners and a final origin. This uses $N_0K_0+2^d+1\le C N_\gamma\Phi_d(\gamma/q)$ insertions. A selector for $k$ and the preceding slot filter retain the origin and precisely this slot. Every membership is $\gamma$-stable. For $j\ne l$, nonnegative inner products and the separation imply
\begin{align}
v_l\cdot[(\rho-a_0(1-W_j))v_j]
&\le\rho\,v_l\cdot v_j\nonumber\\
&\le\rho\bigl(1-\|v_l-v_j\|^2/2\bigr)
\le\rho-2a_0.
\end{align}
The support in direction $v_l$ is therefore exactly $\rho-a_0(1-W_l)$, independently of all other bits.

For completeness, define the fixed $C^2$ cutoffs explicitly. Let $S(t)=0$ for $t\le0$, $S(t)=6t^5-15t^4+10t^3$ for $0<t<1$, and $S(t)=1$ for $t\ge1$. Put
\begin{align}
g(t)&=S(24(t-1/3))S(24(2/3-t)),\nonumber\\
\chi(x)&=\prod_{i=1}^d g(x_i),\qquad
\psi(x)=\prod_{i=1}^d(1-8x_i)_+^3.
\end{align}
Thus $\chi=1$ on $[3/8,5/8]^d$ and vanishes outside $(1/3,2/3)^d$; it is zero throughout the origin component. The bit query is
\begin{equation}
D_{c,l}(x)=\tfrac12-\lambda_0\chi(x)v_l\cdot(x-c)+\sigma\psi(x).
\label{eq:vector-cap-renderer}
\end{equation}
The product rule gives $\|\nabla D\|+\|\nabla^2D\|\le C_d(\lambda_0+\sigma)$ uniformly in $c,v_l$. Moreover $\|\nabla D(x)-\nabla D(y)\|\le C_d(\lambda_0+\sigma)d^{(2-s)/2}\|x-y\|^{s-1}$. Fixing $\lambda_0$ and then $\sigma_0$ sufficiently small satisfies $J,M$ and keeps $D\in[0,1]$. These constants do not depend on $H,n,\gamma,q$.

At the origin the intensity is $h=1/2+\sigma$; the selected slot lies where $\chi=1$, $\psi=0$. Its smallest intensity is
\begin{equation}
b_w=\tfrac12-\lambda_0[\rho-a_0(1-w)].
\end{equation}
All slot intensities lie within $\sigma/8$ of $1/2$ when $\gamma\le c\sigma$. The midpoint $m_w=(h+b_w)/2$ is attained along $te_1$, $0\le t\le1/8$, where $\chi=0$ and $\psi$ continuously decreases from one to zero. Consequently it is the feasible-state oracle, with $r_C\le9\sigma/16$. For any scalar output $z$,
\begin{equation}
\mathcal E_w(z)=(z-m_w)^2+(h-b_w)|z-m_w|.
\end{equation}
The two midpoints differ by $\lambda_0a_0/2=64q$. A wrong nearest-midpoint classification has $|z-m_w|\ge32q$ and hence excess at least $(\sigma/2)32q=16\delta$. Each of the $N_0K_0$ independent bits is thus recoverable with error at most $1/16$. The selector, filter and renderer use only its public address, not its value or the tape. Equation~\eqref{eq:vector-augmented-information}, without supplied prefixes, gives
\begin{equation}
B\ge[1-h_2(1/16)]N_0K_0
\gtrsim_d N_\gamma\Phi_d(\gamma/q).
\end{equation}
This proves Eq.~\eqref{eq:vector-dense-lower}, including public coins, and matches the causal profile upper bound. Additional coordinate assignments cannot weaken the converse because it uses only the original clipped actions.

\paragraph{The same old disagreement and loss profile.}
Fix $\sigma>0$ and consider static insertion streams, with $H=0$ and no coordinate assignments. Place the dense slots in the inner half of $B_\sigma=(1/2,\ldots,1/2)+[-w,w]^d$, where $w=b\sigma$ and fixed $b>0$ is small. Use the $2^d$ corners of $B_\sigma$, the origin, and a final common origin as public anchors. There are $\Theta_d((\sigma/\gamma)^d)$ slots. Taking $\gamma\le w/8$ ensures that every slot filter excludes the box corners with clearance at least $\gamma$.

On $B_\sigma$, $\chi=1$ and $\psi=0$, so every query in Eq.~\eqref{eq:vector-cap-renderer} is affine. Its extrema over every possible unfiltered archive are attained at the public box corners and the origin. Thus all hidden-bit histories have the same old convex hull and the same entire unfiltered loss function for each allowed query. Choosing $b$ small bounds all box intensities by $1/2\pm\sigma/8$. Their midpoint with the origin is attainable along the origin path, so $r_V\le9\sigma/16<\sigma$, while the refined bound also remains $r_C<\sigma$.

For $\gamma=a q^{1/s}$ with fixed small $a>0$ and $s>1$, the matching dense bounds therefore give, for fixed $\sigma$,
\begin{align}
B^{\rm ref,rand}_s\asymp B^{\rm ref,det}_s
&\asymp_\sigma q^{-[(d-1)/2+(d+1)/(2s)]},\nonumber\\
\frac{B^{\rm ref,rand}_s}{B^{\rm unc,det}_s}
&\gtrsim_\sigma q^{-(d-1)(s-1)/(2s)}.
\label{eq:vector-polynomial-correction-cost}
\end{align}
The denominator uses the uncorrected upper bound $B^{\rm unc,det}_s\lesssim_\sigma q^{-d/s}$ under the same radius promise, established in Sec.~\ref{app:smooth-task-memory}. Hence correction has a polynomial cost in every $d\ge2$, even when all old loss profiles agree. The scalar exponent vanishes; its separate within-slot precision argument yields the logarithmic cost instead.
\section{Uncorrected Reference and the Correction Cost}
\label{app:smooth-task-memory}
For fixed $\sigma$ and static insertion streams ($H=0$, without intervening assignments), the finite recurrent hull code gives the uncorrected upper bound used in the correction-cost comparison.

\subsection{An insertion-updatable upper bound}
For compact convex $X\subset B_{R_X}^d$, lift $x$ to $\Phi(x)=(x,\|x\|^2)$ and put $K_U=\operatorname{conv}\Phi(U)$. Lemma~\ref{lem:static-hull-profile} in dimension $p=d+1$ gives a finite insertion code with Hausdorff error $\epsilon$ and $O_d(\epsilon^{-d/2})$ bits, at fixed enclosing radius. Decode $P=\{x\in X:\operatorname{dist}(\Phi(x),\widehat K_U)\le\epsilon\}$. It contains $U$, and any $x\in P$ has a lifted historical mixture $Y$ with
\begin{equation}
\|\mathbb EY-x\|\le2\epsilon,\qquad
|\mathbb E\|Y\|^2-\|x\|^2|\le2\epsilon.
\end{equation}
Writing $\mu=\mathbb EY$, direct expansion gives
\begin{align}
\mathbb E\|Y-x\|^2
&=(\mathbb E\|Y\|^2-\|x\|^2)-2x^\top(\mu-x)\nonumber\\
&\le(2+4R_X)\epsilon=:C_X\epsilon.
\end{align}
Integral Taylor expansion and concavity of $t^{s/2}$ therefore imply, for every pixel of every future frame,
\begin{align}
|f(x)-\mathbb Ef(Y)|
&\le J\|x-\mu\|+(M/s)\mathbb E\|Y-x\|^s\nonumber\\
&\le2J\epsilon+(M/s)(C_X\epsilon)^{s/2}.
\end{align}
The mixture is shared across pixels and frames. Taking $\epsilon=cq^{2/s}$ makes this bound at most $q/4$. The same containing-pool argument as in Sec.~\ref{app:vector-image-bound} gives excess at most $2r_V(q/4)+q^2/16\le\delta$. Hence
\begin{equation}
B_s^{\rm unc,det}(0,\delta,\sigma)\lesssim_d q^{-d/s}.
\label{eq:static-uncorrected-upper}
\end{equation}
At $s=1$, Cauchy--Schwarz instead bounds the image error by $J\sqrt{C_X\epsilon}$ and gives the same $q^{-d}$ endpoint. This uses the classical convex-body entropy scale~\cite{supp-dudley1974entropy,supp-bronshtein1976entropy} with the finite code of Lemma~\ref{lem:static-hull-profile}.

This finite recurrent code supplies the uncorrected upper bound in the correction-cost ratio. Combining it with the refined randomized converse proves the additional memory cost under the same radius promise.
\section{Scalar Specialization}
\label{sec:scalar-correction-sharp}
\subsection{Why a positive margin is necessary}
For arbitrarily large $L$, publish $u(W)$ and then the origin under the fixed renderer $D(x)=\sigma x$, where
\begin{equation}
u(W)=\tfrac12+\sum_{k=1}^L W_k2^{-(k+1)}+2^{-(L+2)}.
\end{equation}
Given the earlier bits, query $C_i=[0,t_i]$ with $t_i=1/2+\sum_{k<i}W_k2^{-(k+1)}+2^{-(i+1)}$. If $W_i=0$, both points survive; otherwise only the origin survives. Their oracle centers are $\sigma u/2$ and zero, both feasible. Classification at $\sigma/6$ incurs excess at least $\sigma^2/36$ for a wrong singleton decision and $7\sigma^2/144$ for a wrong two-state decision. Thus expected excess $\delta\le\sigma^2/512$ permits bit error below $1/8$. Equation~\eqref{eq:vector-augmented-information} forces $B\ge[1-h_2(1/8)]L$ for every $L$. This fixed, smooth renderer therefore has no finite uniform correction memory without clearance, even for two candidates.

\subsection{The all-history scalar rate}
Fix $1<s\le2$, sufficiently small $\sigma$, $0<\delta\le c\sigma^2$ and $0<\gamma\le c\min\{\sigma,q^{1/s}\}$. The vector construction proves the all-$n$ scalar rate without a separate geometric approximation argument. Set $d=1$ in Sec.~\ref{app:vector-upper}. The full bank has two exterior anchors and the other nonzero weight banks disappear. Thus its local field costs $O(\log(2+m))=O(\log(2+\gamma/q))$ bits, with exact updates under all permitted assignments. Occupancy costs $\Psi(2^H/\gamma,n)$; its $\Omega(\min(n,2^H/\gamma))$ term absorbs constant local fields when $\gamma/q$ is bounded.

For the converse, the occupancy argument in Sec.~\ref{app:vector-lower} already gives $\Omega(\Psi(N,n))$ for $n\le cN$, $N=2^H/\gamma$. If $n>cN$, use a fixed small fraction of $N$ occupied slots and pad the remaining insertions by the origin. It gives $\Omega(N)=\Omega(\Psi(N,n))$. For precision at arbitrary $n\ge6$, publicly occupy $a=\Theta(\min(n,N))$ separated slots and choose an independent fine label in each, with the same scalar query and midpoint decoding from that section. These use at most $n$ insertions after including the anchors. Conditional Fano gives $\Omega(a\log(1+\gamma/q))$ when this ratio is large, while occupancy absorbs the other case. The maximum of the two bounds is at least half their sum, proving
\begin{align}
B_s^{\rm ref,rand}(d=1)&\asymp B_s^{\rm ref,det}(d=1)\nonumber\\
&\asymp\Psi(N,n)+\min(n,N)\log_2(1+\gamma/q).
\end{align}
The simpler scalar implementation retains each occupied cell's outward-rounded extrema on denominator $D_r=L2^r/h$, with $L\asymp\max(1,\gamma/q)$. Under $A_b$, integer endpoint $k$ maps exactly to $[k-bD_r/2]_{[0,D_r/2]}$. Insertion takes integer minima/maxima, and clipping merges extrema. Choosing $h\le\gamma/4$, $Jh/L\le q/16$ and $(2M/s)h^s\le q/8$ gives a containing interval pool and the same Taylor/radius bound as the vector code. There are $(L+1)(L+2)/2$ ordered endpoint pairs per occupied cell; ranking the cell subset and these pairs attains the displayed rate for the clipped-action implementation.

At fixed $\sigma$, $H=0$ and $\gamma\asymp q^{1/s}$, publicly occupy $\Theta_\sigma(q^{-1/s})$ slots in a small interval about $1/2$ and add its endpoints and the origin. The endpoints determine every unfiltered extremum for the local quadratic queries; later filters expose the fine labels. The same precision/Fano argument gives $\Theta_\sigma(q^{-1/s}\log(1/q))$ corrected bits under both radius bounds, whereas an uncorrected code uses at most $O_\sigma(q^{-1/s})$ bits. In higher dimensions the radial support choices replace this endpoint logarithm by the polynomial cost proved in Sec.~\ref{app:vector-lower}.
\section{Exact Recovery from Historical Constraints}
\label{app:finite-vector-output}\label{app:finite-loss}
An occupied-cell packet requires $U\subseteq Q$, the union of marked cells, and $U\cap C_j\ne\varnothing$ for every marked cell $C_j$: each cap witnesses a historical member. More generally impose $K\subseteq U$ and quotas $\sum_{x\in U}w_{i,x}\ge q_i$, $w_{i,x}\ge0$. The resulting family $\mathcal H$ is upward closed: compatibility of $U$ implies compatibility of every $U\subseteq U'\subseteq Q$. Its elements are possible complete archives, not individual physical states.

The profile in Sec.~\ref{app:vector-catalogue-caps} supplies the pool and witness caps. The results below apply to its declared family and also allow mandatory states and cardinality quotas.

\paragraph{Exact conditioning of a packet.}
\label{app:family-refinement}
For an upward-closed $\mathcal H$ on a finite pool $Q$, put $Q_{\rm in}=Q\cap C$, $Q_{\rm out}=Q\setminus C$, and $\mathcal H|_C=\{U\cap C:U\in\mathcal H,\ U\cap C\ne\varnothing\}$.
\begin{supplemma}[Completion after a correction]
\label{lem:family-refinement}
For every nonempty $V\subseteq Q_{\rm in}$,
\begin{equation}
V\in\mathcal H|_C\iff V\cup Q_{\rm out}\in\mathcal H.
\label{eq:family-refinement-predicate}
\end{equation}
The refined family is upward closed. Mandatory states become $K\cap C$; caps with a deleted member are removed; a quota with nonnegative weights $w_{i,x}\ge0$ becomes
\begin{equation}
\sum_{x\in V}w_{i,x}\ge\max\left\{0,q_i-\sum_{x\in Q_{\rm out}}w_{i,x}\right\}.
\label{eq:refinement-quota}
\end{equation}
Nonemptiness is imposed separately.
\end{supplemma}
\begin{proof}
Deleted states may have supplied old witnesses, so first adjoin them all. If $V=U\cap C$, then $U\subseteq V\cup Q_{\rm out}$, so upward closure makes the completion feasible. Conversely, choosing that completion as $U$ recovers $V$. Inclusion of refined sets preserves inclusion of completions. The completion already contains every deleted mandatory point and hits every cap with a deleted member. Its quota sum is $\sum_{x\in V}w_{i,x}+\sum_{x\in Q_{\rm out}}w_{i,x}$, proving the update.
\end{proof}
In particular $|U|\ge r$ becomes $|V|\ge\max\{1,r-|Q_{\rm out}|\}$. The exact-risk identity in Sec.~\ref{app:cardinality-witnesses} applies on $Q_{\rm in}$ with the common oracle/output domain. An attaining corrected history extends to a feasible old history by adjoining $Q_{\rm out}$. Intersecting every cap with $C$ gives the wrong family: $U=\{b,c\}$ hits cap $\{a,b\}$, but after $C=\{a,c\}$ its live history $\{c\}$ does not hit the clipped cap $\{a\}$.

\subsection{Transport and merging}
\label{app:monotone-transport}\label{app:general-witness-transport}
To update without replay, characterize each new history by its largest compatible preimage. Let $h(U)=\mathbf1\{U\in\mathcal H\}$, let $A:Q\to X'$ be any deterministic map, let $O\subset X'$ be finite, and put $Q'=A(Q)\cup O$. The exact transported family is
\begin{equation}
\{A(U)\cup O:U\in\mathcal H\}=\{V\subseteq Q':O\subseteq V,\ h(A^{-1}(V))=1\}.
\label{eq:monotone-family-transport}
\end{equation}
For the forward inclusion, $U\subseteq A^{-1}(A(U)\cup O)$ and upward closure imply feasibility. Conversely choose $U=A^{-1}(V)$. It is feasible and
\begin{equation}
A(U)\cup O=(V\cap A(Q))\cup O=V,
\end{equation}
where the last equality uses $O\subseteq V\subseteq Q'$. Preimages preserve inclusion, so the new family is upward closed. Set preimages keep the argument valid for noninjective actions.

Mandatory states become $A(K)\cup O$ and caps become $A(C_j)$. A nonnegative weighted quota transforms exactly as
\begin{align}
\sum_{x\in A^{-1}(V)}w_{i,x}&=\sum_{v\in V}W_{i,v},&
W_{i,v}&=\sum_{x\in Q:A(x)=v}w_{i,x}.
\label{eq:fiber-weight-update}
\end{align}
Thus an old distinct-count bound becomes a fiber-weighted bound. For example, merging two required distinct states into one gives a valid singleton of weight two. Preserving a distinct-count bound of two would incorrectly exclude it. A cap enforcing nonemptiness is included when no other constraint does so. Episode-dependent weights and state maps count toward retained memory.

\subsection{Exact deterministic risk}
\label{app:cardinality-witnesses}
To find worst compatible excess, first minimize the comparator loss. For a nonempty finite output set $Z$, put $F(U)=\min_{y\in Z}\max_{u\in U}\ell(u,y)$. With a finite-valued loss table, define $Q_{y,t}=\{q\in Q:\ell(q,y)\le t\}$ and $a(y)=\min\{t\in\ell(Q,y):Q_{y,t}\in\mathcal H\}$. This minimum exists since $Q\in\mathcal H$. Thus $a(y)$ is the smallest $y$-loss of any compatible history. Fix $x\in Q$ and $y\in Z$. Every feasible history containing $x$ has $y$-loss at least both $\ell(x,y)$ and $a(y)$. The sublevel set at their maximum is feasible, contains $x$, and attains that upper bound. Hence
\begin{align}
\min_{U\in\mathcal H:x\in U}\max_{u\in U}\ell(u,y)&=\max\{\ell(x,y),a(y)\},\\
\min_{U\in\mathcal H:x\in U}F(U)&=\min_{y\in Z}\max\{\ell(x,y),a(y)\}=:b(x).
\end{align}
Here $b(x)$ is the smallest oracle value among histories that contain $x$; the second identity interchanges two finite minima. Exchanging finite maxima then proves
\begin{align}
R(z)&=\max_{U\in\mathcal H}\max_{x\in U}\{\ell(x,z)-F(U)\}\\
&=\max_{x\in Q}\left\{\ell(x,z)-\min_{U\in\mathcal H:x\in U}F(U)\right\}\\
&=\max_{x\in Q}\{\ell(x,z)-b(x)\}.
\label{eq:cardinality-exact-risk}
\end{align}
Every $x\in Q$ occurs because $Q$ is feasible.

\begin{supptheorem}[Witnesses and retained cardinality]
\label{prop:cardinality-witnesses}
For $\mathcal H_r=\{U:K\subseteq U\subseteq Q,\ U\cap C_j\ne\varnothing\ \forall j=1,\ldots,m,\ |U|\ge r\}$ with $K\subseteq Q$, $1\le r\le|Q|$, and nonempty caps $C_j\subseteq Q$, the threshold is
\begin{equation}
a_r(y)=\max\left\{\max_{k\in K}\ell(k,y),\ \max_j\min_{v\in C_j}\ell(v,y),\ \operatorname{os}_r\ell(Q,y)\right\}.
\end{equation}
$\operatorname{os}_r$ is the $r$th smallest loss over distinct states. Empty maxima equal $-\infty$; tied losses retain their multiplicities. Each output risk is attained by a history of size at most $\max\{r,|K|+m+1\}$. All risks take $O(|Z|[|Q|+|K|+\sum_j|C_j|])$ arithmetic operations.
\end{supptheorem}
\begin{proof}
We first test the three constraints on a loss sublevel set, then construct an attaining history. The set must contain $K$, hit all $m$ caps, and contain $r$ distinct states, giving the threshold. To construct a history with smallest oracle while containing $x$, choose a minimizing comparator $y$ in $b(x)$, put $v_j\in\arg\min_{v\in C_j}\ell(v,y)$, and start with $U_0=K\cup\{x,v_1,\ldots,v_m\}$. Every selected loss is at most $\tau=\max\{\ell(x,y),a_r(y)\}$. The order statistic guarantees at least $r$ states below $\tau$, so fill $U_0$ to $r$ if necessary. The resulting history has $y$-loss at most $\tau=b(x)$. Its oracle is at least $b(x)$ by the preceding minimization identity, hence equals $b(x)$. For an $x$ maximizing Eq.~\eqref{eq:cardinality-exact-risk}, this history attains the risk and the stated size bound. Linear-time selection computes the order statistic, and the other operations are scans of the loss table and caps, proving the cost.
\end{proof}
At $r=1$, its order-statistic term is redundant inside the maximum with $\ell(x,y)$. At $r=|Q|$ only the full pool is possible. This also proves that tightening a valid count can only reduce the certificate.

\subsection{Two-pass exact readout}
\label{app:readout-computation}
Let $Q_C=\{u_1,\ldots,u_{N_c}\}$ be the active pool and $\mathcal Z=\{y_1,\ldots,y_{N_o}\}$ the common oracle/output catalogue. Required regions contain $\Lambda_{\rm reg}$ total memberships, counting overlaps, and there are $r_{\rm mand}$ mandatory points. Put $\Lambda=\Lambda_{\rm reg}+r_{\rm mand}$. A minimum distinct-state count $k\ge1$ also enforces nonemptiness. For the loss row $d_y(u)=\ell_G(u,y)$, the feasible-sublevel threshold is
\begin{equation}
a(y)=\max\left\{
\max_j\min_{u\in C_j}d_y(u),\
\max_{u\in K}d_y(u),\
d_{y,(k)}
\right\}.
\label{eq:resource-threshold}
\end{equation}
Empty maxima are omitted. Region minima cost $O(\Lambda_{\rm reg})$, mandatory maxima $O(r_{\rm mand})$, and selecting the $k$th order statistic costs $O(N_c)$. Thus each threshold costs $O(N_c+\Lambda)$.

Partition the output catalogue into blocks $\mathcal B_i$ of at most $b$ rows. Initialize $\beta_0(u)=+\infty$. The first pass updates
\begin{align}
\beta_i(u)&=\min\left\{\beta_{i-1}(u),\
\min_{y\in\mathcal B_i}\max\{d_y(u),a(y)\}\right\},\\
\beta_{\rm final}(u)
&=\min_{y\in\mathcal Z}\max\{d_y(u),a(y)\}=b_C(u).
\end{align}
The equality follows by induction: after block $i$, the minimum is over precisely the first $i$ blocks. Revisit each row in the second pass and compute
\begin{equation}
R_C(y)=\max_{u\in Q_C}\{d_y(u)-\beta_{\rm final}(u)\}.
\end{equation}
This is the exact finite-risk identity in Sec.~F.2. Risk ties use worst pool loss and then output index; Sec.~\ref{app:floor-physical} gives the physical-readout rules. Only a row block, the $N_c$ entries of $\beta$, $N_o$ output risks, and constraint indices are retained:
\begin{align}
\text{arithmetic}&=O(N_o(N_c+\Lambda)),\nonumber\\
\text{auxiliary space}&=O(bN_c+N_c+N_o+\Lambda).
\label{eq:resource-complexity}
\end{align}
A row provider can retain a dense $N_oN_c$ table or recompute rows on both passes. For $T$ frames of at most $P$ components, pairwise loss construction costs $O(TN_oN_cP)$ after frame values are supplied; renderer evaluation and any retained frame features have their own cost. General weighted quotas use the generic sorted-sublevel decoder, adding sorting and quota scans. Constructing explicit attaining histories also incurs their output-size cost.

\begin{table}[t]
\centering\footnotesize\setlength{\tabcolsep}{3pt}
\caption{\textbf{Readout scaling on one CPU thread.} Three four-component frames and eight required regions. Dense-table size is resident MiB; time is median ms over five repeats. Extra is peak traced Python/NumPy allocation above supplied inputs, in MiB. Sorted uses the generic family readout; stream recomputes loss blocks and retains no dense table. Bold compares time and extra allocation within each row.}
\label{tab:readout-scaling}
\begin{tabular}{@{}r|r|rr|rr@{}}\toprule
&Table&\multicolumn{2}{c|}{Sorted}&\multicolumn{2}{c}{Stream}\\
$N_c=N_o$&MiB&ms&Extra&ms&Extra\\\midrule
64&0.03&1.54&\textbf{0.11}&\textbf{0.96}&0.24\\
256&0.50&\textbf{9.94}&1.09&10.31&\textbf{0.58}\\
1024&8.00&\textbf{101.47}&16.13&133.90&\textbf{2.06}\\
2048&32.00&\textbf{377.98}&64.19&401.33&\textbf{4.11}\\\bottomrule
\end{tabular}
\end{table}
The diagnostic varies $N_c,N_o,T$ and region count over all sixteen declared configurations. Complete risk vectors, choices and certificates agree across the generic family readout, the dense decoder and both streamed providers. Loss-table construction is timed separately; for the four sizes in Table~\ref{tab:readout-scaling} it takes $0.43,4.79,66.49,193.68$ ms. A cached provider uses the resident table; at $N_c=N_o=2048$ it takes $23.17$ ms with $1.24$ MiB extra allocation. The uncached provider trades recomputation for lower resident storage. Public renderer weights, provider caches, and model inference memory are separate from the historical packet and these allocation measurements.

\subsection{Disagreement-conditioned truth support}
\label{app:floor-physical}
Let $Q$ be finite, let $\mathcal H$ be a nonempty family of nonempty subsets of $Q$, and let $S_{\rm out}$ be a nonempty compact output set. Assume $\ell(u,x)$ is continuous in $x$ for every $u$. Write $F(U)=\min_{x\in S_{\rm out}}\max_{u\in U}\ell(u,x)$ and
\begin{equation}
b(u)=\min_{U\in\mathcal H:u\in U}F(U),
\end{equation}
with $b(u)=+\infty$ if no such history exists. The minima are attained because there are finitely many histories and their losses are continuous on a compact set. For an upward-closed family, the sublevel identity computes $b$ without enumerating histories.

\begin{supptheorem}[Truth support and physical risk]
For public $f\ge0$, put $\mathcal H_f=\{U\in\mathcal H:F(U)\le f\}$ and assume it is nonempty. Then
\begin{align}
Q_f&:=\{u\in Q:b(u)\le f\}=\bigcup_{U\in\mathcal H_f}U,\label{eq:supp-floor-support}\\
R_f(x)&=\max_{U\in\mathcal H_f}[L(U,x)-F(U)]\nonumber\\
&=\max_{u\in Q_f}[\ell(u,x)-b(u)],\label{eq:supp-floor-risk}\\
P_f(x)&=\max_{U\in\mathcal H_f}\max_{u\in U}\ell(u,x)
=\max_{u\in Q_f}\ell(u,x).\label{eq:supp-floor-physical}
\end{align}
The family $\mathcal H_f$ need not be upward closed. If a sound current set $S$ also contains the true state, replace $Q_f$ in Eq.~\eqref{eq:supp-floor-physical} by $Q_f\cap S$.
\end{supptheorem}
\begin{proof}
If $u\in U\in\mathcal H_f$, minimization gives $b(u)\le F(U)\le f$, proving one inclusion in Eq.~\eqref{eq:supp-floor-support}. If $b(u)\le f$, an attaining history $U_u\ni u$ has $F(U_u)=b(u)$ and belongs to $\mathcal H_f$, proving the reverse inclusion. For every eligible $u$, the minimum of $F(U)$ over $U\in\mathcal H_f$ containing it is therefore still $b(u)$. Interchanging the two finite maxima gives
\begin{align}
R_f(x)&=\max_{U\in\mathcal H_f}\max_{u\in U}[\ell(u,x)-F(U)]\nonumber\\
&=\max_{u\in Q_f}\left[\ell(u,x)-\min_{U\in\mathcal H_f:u\in U}F(U)\right],
\end{align}
which is Eq.~\eqref{eq:supp-floor-risk}. Maximizing over the union proves Eq.~\eqref{eq:supp-floor-physical}. Its maximizing state has the witness $U_u$, so the bound is attained by a compatible history and a covered truth. Adding $u\in S$ changes only the allowed truth states and preserves the same witness argument. It does not require every alternative in $U$ to belong to $S$.
\end{proof}

For an actual $U\in\mathcal H_f$ and $x_{\rm true}\in U$, Eq.~\eqref{eq:supp-floor-physical} gives $\ell(x_{\rm true},x)\le P_f(x)$. Also $b(u)\le f$ on $Q_f$, hence $P_f(x)\le f+R_f(x)$. With $\ell(u,x)=\max_j\|G_j(u)-G_j(x)\|_{P_j}^2$ and $\max_j\|I_j-G_j(x_{\rm true})\|_{P_j}\le\varepsilon$, the triangle inequality yields
\begin{equation}
\max_j\|I_j-G_j(x)\|_{P_j}\le\varepsilon+\sqrt{P_f(x)}.
\end{equation}
At $\varepsilon=0$, $P_f$ is the exact worst compatible true-image loss. For nonzero calibration error the displayed upper bound need not be attained inside the image cube. If $\mathcal H_f$ is empty, no prediction is authorized by this promise.

\paragraph{Physical readout with historical tie resolution.} Let $p_* =\min_{x\in S_{\rm out}}P_f(x)$ and choose
\begin{equation}
\widehat x\in\arg\min_{x:P_f(x)=p_*}R_f(x).
\end{equation}
Continuity and compactness give both minima. This rule attains the optimal physical worst-case risk and has the smallest historical excess certificate among physical optima. Finite tests break remaining ties by mean pool loss and then catalogue index. A pool control minimizes physical pool loss, then the same mean and index; the retained-history profile is the sole extra information used for secondary selection.

\paragraph{Independent witnesses.} For a required cap $A$, put $a_A(y)=\min_{v\in A}\ell(v,y)$. A mandatory state is a singleton cap. The independent lower profile is
\begin{align}
b_{\rm ind}(u)=\max\bigl\{&\min_y\ell(u,y),\nonumber\\
&\max_A\min_y\max\{\ell(u,y),a_A(y)\}\bigr\}.
\end{align}
The min--max inequality gives $b_{\rm ind}(u)\le b(u)$. Thus $Q_f\subseteq Q_{{\rm ind},f}\subseteq Q$, and their physical minimax values are ordered in the same direction. More directly, the exact physical decoder has no greater risk on the common possible-truth set $Q_f$ than any decoder chosen by the weaker checks. 

Strict inequality is realizable with identity two-pixel images. Take $a=(0,0)$, $b=(1,0)$ and $v=(1/2,\sqrt3/2)$, output the whole triangle, and use $\ell(u,x)=\|u-x\|^2/2$. The packet requires $a,b$ and permits $v$. The two histories have floors $F(\{a,b\})=1/8$ and $F(\{a,b,v\})=1/6$. Hence $b(a)=b(b)=1/8$, $b(v)=1/6$, while every independent pair has radius squared $1/8$. At $f=1/8$, exact support is $\{a,b\}$ and its midpoint incurs worst true loss $1/8$. Independent and pool minimax retain the triangle; its circumcenter incurs $1/6$ even against the same two truths. This improvement uses one shared comparator for all witnesses, rather than separate pairwise centers.

\paragraph{Every radius from one profile.} For $N_o$ finite outputs and $N_c$ candidates, sort $b(u)$ once. Process equal values together. Each output row supplies prefix maxima of $\ell(u,x)$ and $\ell(u,x)-b(u)$, and prefix sums for mean ties. Scan the outputs with linear minimum-and-tie masks to retain the best objective at each critical budget. This costs $O(N_oN_c+N_c\log N_c)$ time and $O(N_c)$ additional storage beyond the streamed loss row and supplied data. It includes every threshold where support changes; no history enumeration or persistent radius bank is needed.

\paragraph{When excess and pool decisions coincide.} Let $\Delta=\max_ub(u)-\min_ub(u)$ on a common nonempty pool and let $x_0$ minimize $P(x)=\max_u\ell(u,x)$. Since $P(x)-\max b\le R(x)\le P(x)-\min b$, an excess minimizer $x_* $ satisfies
\begin{equation}
0\le R(x_0)-R(x_*)\le\Delta.
\end{equation}
If $b$ is constant the argmin sets coincide. If $x_0$ is unique and every other output has pool-loss gap larger than $\Delta$, it remains the unique excess minimizer. Equal argmin sets need the same deterministic tie rule to imply the same selected index.
\section{Joint-State Photographic Continuation}
\label{app:joint-photo}
\paragraph{Coupled visibility and common-state evolution.} The twenty cyclic pairs $(1,2),\ldots,(20,1)$ use every COIL-20 object and all seventy-two starting views. Each object has phases $0,120^\circ,240^\circ$; index $3a+b$ denotes phase pair $(a,b)$. The original $128\times128$ photos overlap by $64$ pixels on a $128\times192$ canvas. Nonzero pixels supply foreground alpha; B remains the front layer after exchanges. Normalized MSE uses the full canvas.

The three frames rotate A, exchange positions while rotating B, then rotate A again, all from one initial phase pair. Of $4{,}320$ frame-loss matrices, $4{,}173$ are nonseparable across phase coordinates. A disjoint control uses side-by-side photos and their average MSE under the same state, action and observation protocol.

\paragraph{Complete histories and finite encoders.} All $511$ nonempty histories and seven observations---the full set and each exact phase of either object---give $5{,}150{,}880$ nominal queries: $4{,}606{,}560$ nonempty corrections and $544{,}320$ empty cases. Grouping each phase index as $\{0,1\}$ or $\{2\}$ gives four product cells, whose occupancies form the four-bit joint packet; the four-bit point packet retains the smallest public index or an empty symbol. Both insertion updates are causal. Every point-pool control uses the rank-implied absence of smaller indices. Enumeration verifies both corrected families. The full-history reference uses nine bits.

Physical readout orders states by $(P_f,R_f,\text{mean pool loss},\text{index})$; pool minimax uses its physical objective followed by the same last two rules. Excess readout orders by $(R_f,P_f,\text{mean pool loss},\text{index})$. Independent-witness readout replaces $b_C$ by the separate-cap profile in Sec.~\ref{app:floor-physical}. These rules and nine public budgets $f\in\{0,1/400,1/200,1/100,1/50,1/25,2/25,4/25,1\}$ are fixed across all source pairs. At each budget every history with $F^C\le f$ is included; remaining histories are counted independently of method outcomes. At $f=1$, all nonempty queries are admissible.

\paragraph{Image quantities.} $L,R_f,P_f$ are defined in the main paper. Common $P_f$ evaluates every control on the identical exact truth support $Q_f$. Target MSE averages over uniform candidate truths within each actual history; singletons specify an unambiguous trajectory. The full-history physical loss equals its oracle floor.

\begin{table*}[t]
\centering\small\setlength{\tabcolsep}{4.1pt}
\caption{\textbf{Complete joint-state photographic results at $f=1$.} Every entry uses all $4{,}606{,}560$ nonempty corrected queries. Losses are scaled by $10^3$ (lower is better). $L$ is actual historical loss, $R$ exact family excess, $P$ common possible-truth loss, and $e$ mean target loss. Bold compares four-bit methods, with the nine-bit reference separate.}
\label{tab:joint-photo-full}
\begin{tabular}{@{}lrrrrrrrr@{}}\toprule
&\multicolumn{4}{c}{Occluding canvas}&\multicolumn{4}{c}{Disjoint control}\\
Method & $L$ & $R$ & $P$ & $e$ & $L$ & $R$ & $P$ & $e$\\\midrule
Joint: pool & 49.275 & 38.946 & \textbf{56.313} & 35.599 & 44.037 & 35.465 & \textbf{51.208} & 31.436\\
Joint: excess & 49.287 & 38.836 & 56.398 & 35.513 & 44.034 & 35.463 & 51.210 & 31.431\\
Joint: physical & 49.278 & 38.937 & \textbf{56.313} & 35.610 & 44.036 & 35.464 & \textbf{51.208} & 31.436\\
Joint: independent & 49.278 & 38.938 & \textbf{56.313} & 35.610 & 44.036 & 35.464 & \textbf{51.208} & 31.436\\
Point: native & 48.340 & 33.337 & 60.060 & \textbf{31.745} & 43.124 & 29.666 & 53.620 & \textbf{28.160}\\
Point: pool & 49.604 & 45.803 & 58.214 & 35.767 & 43.719 & 37.762 & 53.026 & 30.299\\
Point: excess & \textbf{48.075} & \textbf{33.113} & 59.761 & 31.904 & 42.840 & \textbf{29.619} & 53.418 & 28.384\\
Point: physical & 49.415 & 44.943 & 58.214 & 35.515 & \textbf{42.739} & 33.694 & 53.026 & 28.965\\
Point: independent & 49.415 & 44.943 & 58.214 & 35.515 & \textbf{42.739} & 33.694 & 53.026 & 28.965\\
\midrule
Full-history oracle & \textbf{38.233} & \textbf{0.000} & \textbf{38.233} & \textbf{24.164} & \textbf{34.614} & \textbf{0.000} & \textbf{34.614} & \textbf{19.328}\\
\bottomrule\end{tabular}
\end{table*}

\begin{table}[t]
\centering\small\setlength{\tabcolsep}{4pt}
\caption{\textbf{All fixed budgets and admissible histories.} Both encoders and every readout use these same cohorts; $4{,}606{,}560$ is the common nonempty total. Remaining histories are outside the floor promise.}
\label{tab:all-floor-cohorts}
\begin{tabular}{@{}lrr@{}}\toprule
$f$ & Occluding & Disjoint\\\midrule
$0$ & 1,671,840 & 1,671,840\\
$1/400$ & 1,684,710 & 1,678,860\\
$1/200$ & 1,781,937 & 1,878,210\\
$1/100$ & 2,039,337 & 2,157,840\\
$1/50$ & 2,316,756 & 2,362,557\\
$1/25$ & 2,752,071 & 2,727,069\\
$2/25$ & 3,742,308 & 3,878,118\\
$4/25$ & 4,557,549 & 4,565,910\\
$1$ & 4,606,560 & 4,606,560\\
\bottomrule\end{tabular}
\end{table}

\paragraph{Fixed-budget outcomes.} Table~\ref{tab:joint-photo-full} retains all ten readouts for both renderers. Point-packet physical tie resolution lowers historical and mean target loss against pool minimax at the same physical bound. Native point retention has lower target error and weaker physical risk. Joint occupancy has the strongest equal-bit physical bound; its physical tie rule slightly increases target loss at $f=1$. Pair-reweighting intervals describe catalogue variation: neighboring pairs share objects.

\paragraph{All critical disagreement levels.} All exact/independent profile breakpoints, including $0,1$, give $198{,}720$ contexts and $531{,}723$ levels: $36{,}421$ infeasible and $495{,}302$ admissible packet/budget pairs. Exact witnesses shrink truth support in $5{,}610$ pairs and strictly lower common physical loss in $2{,}839$; all others tie. Every strict gain uses joint occupancy; its $285{,}793$ admissible pairs supply the main figure's denominator. Point packets require one witness and give no separation. Disjoint images give none in $465{,}631$ admissible pairs. These are complete structural spectra; the nine fixed budgets evaluate performance separately. Spectrum frequencies are distinct from natural episode frequencies.

\begin{figure}[t]
\centering\includegraphics[width=\columnwidth]{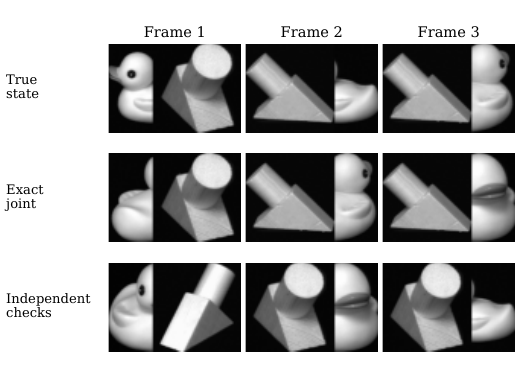}
\caption{\textbf{Joint witnesses change a visual decision.} Under the given disagreement bound, joint constraints exclude phase pair $(1,0)$, which independent checks retain. Exact and independent readouts choose $(0,2)$ and $(1,1)$. Each row evolves from one initial phase pair; the top row is the compatible truth $(2,2)$. All frames show the full canvas.}
\label{fig:joint-occlusion-case}
\end{figure}

Figure~\ref{fig:joint-occlusion-case} is the first strict risk difference in protocol order: pair $(1,2)$, start~1, unrestricted filter and $f=0.1699706$. All four coarse cells are occupied. Truth $(2,2)$ belongs to history $\{1,2,6,8\}$ with oracle loss $f$ and attains the independent output's worst-truth loss. The common risk bounds are $0.1816973/0.1817578$; errors on this trajectory are $0.0586183/0.1817578$.
\section{Encoder Precision and Causal Transport}
\label{app:smooth-photo-endpoints}
All $20\times72$ COIL starts define $G(x)=(1-x)^2I_a+2x(1-x)I_b+x^2I_c$ from photographs at offsets $0,20,40$ degrees. Publish sixteen rational states per $h=1/64$ cell within $h/8$ of its center, then $0$ ($H=0,n=1025$). The six filters $C=[0,5/16]\cup[i/64,(i+1)/64]$, $i\in\{24,28,32,36,40,44\}$, have clearance $1/256$; alignment keeps decoded intervals in $C$. Shared outputs comprise $257$ uniform origin-interval points and $17$ selected-cell points. Integer Gram numerators, stationary roots and endpoints give interval maxima.

Pixelwise interpolation error is at most $M(x-a)(b-x)/2\le Mh^2/8$, also bounding normalized image error. For $L$ subdivisions, $\epsilon=(64L)^{-1}$, $J_*=2$, $M_*=4$, and
\begin{equation}
\eta_*(L)=\frac{2}{64L}+\frac{1}{8192},\qquad
\delta_*(L)=2\eta_*(L)+\eta_*(L)^2.
\label{eq:precision-common-target}
\end{equation}
Nearest rounding on the grid of denominator $D=\lceil2/\eta_*\rceil$ gives two-sided radius perturbation $J_*/D\le\eta_*$. Both codes therefore guarantee the same $\mathcal E\le\delta_*$ under $r_C\le1$. At $L=64$, $D=3277$, $\eta_*=5/8192$, and $\delta_*=81945/67108864$. The test compares fixed codes at the common analytic target.

\begin{table*}[t]
\centering\small\setlength{\tabcolsep}{3pt}
\caption{\textbf{Complete endpoint-precision ablation.} Each row uses all $8640$ queries. Bits include byte padding; excesses are scaled by $10^6$, targets by $10^3$. Bold compares matched-target codes. Full-history excess is zero; occupied cells use $64$ bits with mean/max excess $92.546/549.172$. Only occupied cells violate targets.}
\label{tab:smooth-photo-endpoints}
\begin{tabular}{@{}rr|rrr|rrr|r@{}}\toprule
& &\multicolumn{3}{c|}{Endpoint intervals}&\multicolumn{3}{c|}{Nearest fine subset}&Occupied cells\\
$L$ & $\delta_*$ & Bits & Mean & Max. & Bits & Mean & Max. & Violations\\\midrule
64 & 1.221076 & \textbf{712} & \textbf{0.014349} & \textbf{7.393842} & 2936 & 0.044487 & 14.092218 & 0/8640\\
128 & 0.732556 & \textbf{840} & 0.014349 & 7.393842 & 3800 & \textbf{0.007484} & \textbf{3.933667} & 0/8640\\
256 & 0.488341 & \textbf{968} & \textbf{0.001964} & \textbf{1.738258} & 4456 & 0.007681 & 3.572665 & 5/8640\\
512 & 0.366244 & \textbf{1096} & \textbf{0.000406} & \textbf{0.591101} & 4904 & 0.002235 & 1.607772 & 101/8640\\
1024 & 0.305199 & \textbf{1224} & \textbf{0.000000} & \textbf{0.000000} & 5184 & 0.000949 & 2.540866 & 290/8640\\
\bottomrule\end{tabular}\end{table*}
The fine grid is more accurate at $L=128$. Curvature contributes $20\%$--$80\%$ of the uniform $\eta_*$; empirical endpoint excess nevertheless reaches zero at $L=1024$, whose grid aligns with insertion denominator $65536$. The analytic curvature term remains positive at this grid-aligned setting.

\subsection{Vector code diagnostic}
The twelve configurations cross $d=1,2,3$ with $m=1,2,4,8$. Three occupied cells are encoded, including framing, count, subset rank and every padded signed field. Decoded profiles are compared with literal minima before and after copies, swaps, reflections, resets, merges and clipped expansion. The ablation stores integer fields separately. At $m=8$, cell profiles cost $22$, $2{,}552$ and $577{,}776$ bits in dimensions one, two and three. Packet capacity and byte padding are accounted separately.

\subsection{Causal scalar diagnostic}
Set $G(x)=x$, $J=M=1$, $\sigma=1$, $\delta=q$, with
\begin{align}
q&\in\{2^{-6},2^{-8},2^{-10},2^{-12},2^{-16}\},\nonumber\\
\gamma&=\sqrt q/a,\quad a\in\{8,16,32\},\quad h=\gamma/4,\nonumber\\
H&\in\{0,2,4\},\quad n\in\{8,32,128,512\}.
\end{align}
Use the smallest power-of-two $L$ with $h/L\le q/16$; all settings also obey $h^2\le q/8$. These parameters provide a nonasymptotic test of the integer code. Exact alphabet counts cover all $180$ settings. At $n=8,32$, eight fixed dyadic streams per configuration publish exactly $n$ points, interleaving insertions and a selector word of $H$ clipped actions. Old candidates include a selected parent's interior points and other parents' points; the last common insertion is $0$. Two queries retain a local band, or that band plus $[0,5/16]$. Every actual point is checked against the promised $\gamma$ clearance, including clipped domain endpoints.

Compare the actual endpoint packet, an occupied-subset packet, and late endpoint precision. The last method first encodes with $L=1$, then re-expresses its stored outer intervals at the final $L$, without replaying old points. All share an available ceiling; native alphabets and bytes differ. For each query, literal evaluation finds $a=\min V_C$, $b=\max V_C$, $\mu=(a+b)/2$, and minimizes
\begin{equation}
\max_{u\in V_C}(u-x)^2=\bigl((b-a)/2+|x-\mu|\bigr)^2,\quad x\in C.
\end{equation}
Projection onto each allowed interval and endpoint comparison give its exact oracle. Every encoder prefix is independently compared with transported literal extrema. Table~\ref{tab:causal-diagnostic} summarizes all $1{,}440$ queries per method; every query satisfies the stated clearance, and no fallback occurs. Finest-grid endpoint errors vanish through grid alignment.

\begin{table}[t]
\centering\footnotesize\setlength{\tabcolsep}{3pt}
\caption{\textbf{Causal precision.} Maximum $\mathcal E/\delta$ and target-violation counts (out of $288$ queries at each $q$). E/O/L denote endpoint, occupancy, and late precision. Bold marks best values, including ties.}
\label{tab:causal-diagnostic}
\begin{tabular}{@{}r|rrr|rrr@{}}\toprule
$-\log_2q$ & \multicolumn{3}{c|}{Maximum $\mathcal E/\delta$} & \multicolumn{3}{c}{Violations}\\
& E & O & L & E & O & L\\\midrule
6 & \textbf{0.0145} & 0.0570 & 0.0570 & \textbf{0} & \textbf{0} & \textbf{0}\\
8 & \textbf{0.0137} & 0.1156 & 0.1156 & \textbf{0} & \textbf{0} & \textbf{0}\\
10 & \textbf{0.0117} & 0.2328 & 0.2328 & \textbf{0} & \textbf{0} & \textbf{0}\\
12 & \textbf{0.0078} & 0.4672 & 0.4672 & \textbf{0} & \textbf{0} & \textbf{0}\\
16 & \textbf{0.0000} & 1.8734 & 1.8734 & \textbf{0} & 18 & 18\\
\bottomrule\end{tabular}
\end{table}

\section{Photographic Decisions and Correction}
\label{app:experiments}
COIL-20 supplies twenty objects, each photographed at seventy-two calibrated rotations~\cite{supp-nene1996coil}. We use the full object/view catalogue to isolate corrected historical information. The four-bit cell packet is evaluated with excess readout.

\subsection{Exhaustive corrected photographic histories}
For each of the $20$ objects and $72$ starting views $v$, use eight public states spaced by $45^\circ$. The three future frames are actual source photographs at $v+9(x+j)$ modulo $72$, $j=1,2,3$. Squared pixel loss uses all $128^2$ pixels, normalized by $255^2$. Losses are evaluated directly on source photographs. A current camera reports one of the circular state intervals of lengths $2,4,6$ at every starting state, or the full set: $25$ filters.

Each of the $255$ nonempty histories is reachable through eight insertions (repeating a member to pad), a fixed $90^\circ$ rotation, and a final identity action. Four cell indicators retain adjacent state pairs. Their actual encoder fiber is exactly all subsets hitting each marked cell. Completion after correction remains exact, including partial-cell deletion. For the four-bit point channel, storing the minimum original index $m$ has fiber $m\in U\subseteq\{m,\ldots,7\}$. After the shared permutation $T$, put $Q_m=T(\{m,\ldots,7\})$, $z_m=T(m)$. Its corrected fiber is
\begin{equation}
\{\varnothing\ne W\subseteq Q_m\cap C:z_m\in C\Rightarrow z_m\in W\}.
\end{equation}
Necessity is immediate. If $z_m\notin C$, $W\cup\{z_m\}$ completes any such $W$; otherwise $W$ already does. All completions have at most eight states and are reachable by the same padded schedule. Thus neither channel introduces unreachable-history slack. Literal enumeration verifies all $375$ cell and $200$ point families under this shared rotation.

Table~\ref{tab:photographic-fibers} reports all $8{,}246{,}880$ queries with nonempty corrected histories; $933{,}120$ empty cases are explicitly recorded with undefined excess. The full archive is an eight-bit reference. The point channel is evaluated with both its native decision and the exact decision for its reachable family. Exact risk computations match literal history enumeration. The common-state price is positive in $42.25\%$ of queries, with mean $0.00413$.

\begin{table}[t]
\centering\small\setlength{\tabcolsep}{4pt}
\caption{\textbf{Complete finite-photo outcomes.} All losses are scaled by $10^3$. $R$ is true-fiber worst excess, $\mathcal E$ actual excess, and $L=F+\mathcal E$ historical image loss. Bold compares equal-bit methods and, separately, the full reference.}
\label{tab:photographic-fibers}
\begin{tabular}{@{}lr|rrr@{}}\toprule
Method & Bits & $R\downarrow$ & $\mathcal E\downarrow$ & $L\downarrow$\\\midrule
CAM, excess & 4 & 52.176 & 13.616 & 74.634\\
Same-packet outer & 4 & 53.535 & 13.583 & 74.601\\
Point, direct & 4 & 45.844 & 13.130 & 74.148\\
Point, exact family & 4 & \textbf{45.476} & \textbf{12.762} & \textbf{73.780}\\\midrule
Full oracle & 8 & \textbf{0.000} & \textbf{0.000} & \textbf{61.018}\\
\bottomrule\end{tabular}
\end{table}

For the cell channel, exact and outer readout differ on $1{,}220{,}994/8{,}246{,}880$ queries; every changed decision strictly lowers true-fiber worst risk. Across all histories, actual loss is smaller in $310{,}151$ cases, equal in $7{,}304{,}397$, and larger in $632{,}332$. The point channel has lower overall mean loss. The coverage sweep uses every threshold $k/2000$, $k=0,\ldots,100$.

For all $2{,}499{,}840$ single-survivor queries, the oracle floor is zero and historical loss is MSE to one unambiguous trajectory. Excess and outer mean MSEs are $0.0374204/0.0377752$; seventeen object means favor excess readout. Whole-object $95\%$ intervals use $20{,}000$ paired resamples. Their mean difference, scaled by $10^3$, is $-0.355$ [$-0.534,-0.188$]. For all $5{,}747{,}040$ multiple-survivor queries it is $0.202$ [$0.120,0.288$]; the complete-cohort difference is $0.033$ [$-0.043,0.109$].

The stronger point-pool control uses every rank-implied exclusion and resolves physical ties by mean pool loss. At $f=1$, exact physical tie resolution reduces its mean historical loss from $0.0761000$ to $0.0743265$ and target MSE from $0.0556320$ to $0.0533975$, at the identical physical bound $0.0889981$. Native point retention has lower target MSE $0.0479940$ and weaker physical bound $0.0922988$.

The first lexicographic single-object decision disagreement has pool $\{2,3,6\}$ and requires one of $\{2,3\}$. Excess and outer readouts select states $3$ and $4$. The six compatible histories $\{2\},\{3\},\{2,3\},\{2,6\},\{3,6\},\{2,3,6\}$ explain both the lower worst excess and the history-dependent loss tradeoff.

\section{Neural Continuation: Protocol and Metric}
\label{app:neural-continuation}
\begin{table*}[t]
\centering\small\setlength{\tabcolsep}{4pt}
\caption{\textbf{Complete neural continuation.} All $2{,}048$ sequences per method; $30{,}720$ in total. Loss is $10^3e_{\rm img}$ and accuracy is the all-frame joint percentage; P/E denote preserving/erasing requests.}
\label{tab:neural-all}
\begin{tabular}{@{}lrrrrrrrrr@{}}\toprule
& & \multicolumn{2}{c}{Full/P} & \multicolumn{2}{c}{Upper/P} & \multicolumn{2}{c}{Full/E} & \multicolumn{2}{c}{Upper/E}\\
Method & Bits & Loss$\downarrow$ & Acc.$\uparrow$ & Loss$\downarrow$ & Acc.$\uparrow$ & Loss$\downarrow$ & Acc.$\uparrow$ & Loss$\downarrow$ & Acc.$\uparrow$\\\midrule
CAM, physical & 64 & 71.71 & 84.8 & 69.43 & 45.1 & 63.94 & 86.3 & 64.88 & \textbf{87.1}\\
CAM, excess & 64 & 71.68 & 86.1 & 71.37 & 55.3 & 63.88 & \textbf{87.3} & 64.72 & 86.9\\
Sound-pool minimax & 64 & 71.83 & 84.8 & 69.32 & 45.1 & 64.59 & 86.3 & 64.46 & \textbf{87.1}\\
Outer-pool minimax & 64 & 71.68 & 86.1 & 71.27 & 55.3 & 63.88 & \textbf{87.3} & 64.88 & \textbf{87.1}\\
Unconditioned & 64 & 71.65 & 84.6 & 77.21 & 6.8 & 61.13 & 86.3 & 64.72 & 86.9\\
Recursive joint filter & 64 & 71.83 & 84.8 & 71.33 & 45.1 & 64.59 & 86.3 & 65.30 & 86.9\\
Ranked retention & 64 & 70.58 & 85.9 & 65.84 & \textbf{56.1} & 65.13 & 86.9 & 64.97 & 86.9\\
Factorized marginals & 32 & 74.08 & 84.4 & 67.71 & 16.6 & 62.53 & 86.3 & 65.30 & 86.9\\
Latest relative view & 31 & 72.48 & \textbf{86.3} & 73.91 & 21.7 & 66.47 & \textbf{87.3} & 64.43 & \textbf{87.1}\\
\midrule
Current-set minimax & 0 & 71.91 & 84.4 & 55.41 & 12.9 & \textbf{60.53} & 86.3 & \textbf{64.39} & 86.9\\
Current point & 0 & \textbf{70.04} & 84.6 & \textbf{55.14} & 12.9 & 60.95 & 86.3 & 64.66 & 86.9\\
Current + scene prior & 0 & 73.53 & 85.5 & 75.33 & 4.7 & 63.72 & 86.9 & 64.66 & 86.9\\
\midrule
Exact interval program & 48 & 71.72 & 84.6 & 70.84 & 38.5 & 61.02 & 86.5 & 64.88 & \textbf{87.1}\\
Full fine archive & 256 & 71.72 & 84.6 & 70.84 & 38.5 & 61.02 & 86.5 & 64.88 & \textbf{87.1}\\
Full recursive belief & 256 & 72.48 & \textbf{86.3} & 74.01 & 20.3 & 66.47 & \textbf{87.3} & 65.30 & 86.9\\
\bottomrule\end{tabular}
\end{table*}

\paragraph{Metric and parser failures.} Let $\mathcal Y=[0,1]\cup\{\bot\}$, where $\bot$ denotes a missing or unparseable object. Define
\begin{equation}
d_\bot(a,b)=\begin{cases}|a-b|,&a,b\in[0,1],\\0,&a=b=\bot,\\1,&\text{exactly one of }a,b\text{ is }\bot.\end{cases}
\label{eq:neural-bottom-distance}
\end{equation}
For two-object states and $T$-frame sequences, use
\begin{align}
d_2(u,v)&=\left[\tfrac12\sum_{i\in\{A,B\}}d_\bot(u_i,v_i)^2\right]^{1/2},\nonumber\\
d_T(\mathbf u,\mathbf v)&=\max_{1\le j\le T}d_2(u_j,v_j).
\label{eq:neural-sequence-metric}
\end{align}
\begin{supplemma}[Metric and generation-error bridge]
\label{lem:neural-metric-bridge}
Both $d_2,d_T$ are metrics. Let $\mathbf y,\mathbf g,\mathbf z$ be the image-parsed generation, commanded trajectory, and prescribed target trajectory. For
\begin{equation}
e_{\rm img}=d_T(\mathbf y,\mathbf z)^2,\quad e_{\rm state}=d_T(\mathbf g,\mathbf z)^2,\quad\rho=d_T(\mathbf y,\mathbf g),
\end{equation}
we have
\begin{equation}
\bigl|\sqrt{e_{\rm img}}-\sqrt{e_{\rm state}}\bigr|\le\rho,
\qquad e_{\rm img}\le(\sqrt{e_{\rm state}}+\rho)^2.
\label{eq:neural-generation-bridge}
\end{equation}
\end{supplemma}
\begin{proof}
On three numerical points, $d_\bot$ is absolute distance. If $b=\bot$ and $a,c$ are numerical, $|a-c|\le1\le2$. By symmetry, the remaining nonzero case has $a=\bot,c\ne\bot$: numerical $b$ gives $1\le1+|b-c|$; $b=\bot$ gives $1\le0+1$. If $a=c=\bot$, the left side vanishes. The other axioms are direct.

For the product, put $a_i=d_\bot(u_i,v_i)$ and $b_i=d_\bot(v_i,w_i)$. Coordinatewise triangles and Cauchy--Schwarz give
\begin{align}
2d_2(u,w)^2&\le\|a+b\|_2^2=\|a\|_2^2+2a^\top b+\|b\|_2^2\nonumber\\
&\le(\|a\|_2+\|b\|_2)^2.
\end{align}
Taking square roots yields $d_2(u,w)\le d_2(u,v)+d_2(v,w)$. Therefore
\begin{align}
d_T(\mathbf u,\mathbf w)&\le\max_j\{d_2(u_j,v_j)+d_2(v_j,w_j)\}\nonumber\\
&\le d_T(\mathbf u,\mathbf v)+d_T(\mathbf v,\mathbf w).
\end{align}
Other axioms are coordinatewise. Using intermediate $\mathbf g$, then $\mathbf y$, gives
\begin{equation}
\sqrt{e_{\rm img}}\le\rho+\sqrt{e_{\rm state}},\qquad
\sqrt{e_{\rm state}}\le\rho+\sqrt{e_{\rm img}}.
\end{equation}
Subtract, then square the first inequality.
\end{proof}
The measured post-generation $\rho$ decomposes selection and control error; pre-generation guarantees use the calibration hypothesis in the main paper. Erasing actions induce pullback pseudometrics, and the square defines their associated loss.

\begin{table}[H]
\centering\footnotesize\setlength{\tabcolsep}{2.8pt}
\caption{\textbf{Neural state selection and colour-only image loss.} Entries are $10^3e_{\rm state}/10^3e_{\rm HSV}$ for all $512$ sequences in each condition. P/E denote preserving/erasing requests.}
\label{tab:neural-secondary}
\begin{tabular}{@{}lrrrr@{}}\toprule
Method & Full/P & Upper/P & Full/E & Upper/E\\\midrule
CAM, physical & 0.21/2.25 & 1.97/4.93 & 0.43/0.47 & 0.39/0.43\\
CAM, excess & 0.18/2.21 & 1.83/5.71 & 0.36/0.41 & 0.23/0.27\\
Sound-pool & 0.33/2.37 & 1.87/4.83 & 0.11/0.15 & 0.95/0.99\\
Outer-pool & 0.18/2.21 & \textbf{1.73}/5.61 & 0.36/0.41 & 0.39/0.43\\
Unconditioned & 4.66/6.73 & 38.52/39.17 & 0.54/0.59 & 0.23/0.27\\
Recursive & 0.33/2.37 & 1.93/4.07 & 0.11/0.15 & 0.80/0.85\\
Ranked retention & 1.04/3.08 & 5.16/5.38 & 0.64/0.68 & 1.45/1.49\\
Factorized & 1.58/3.66 & 15.80/16.42 & 1.92/1.99 & 0.80/0.85\\
Relative view & \textbf{0.02}/\textbf{2.04} & 2.68/\textbf{2.87} & \textbf{0.03}/\textbf{0.07} & 0.92/0.96\\
Current minimax & 1.39/3.45 & 21.46/22.10 & 0.92/0.96 & 0.88/0.91\\
Current point & 1.48/3.53 & 21.20/21.84 & 0.36/0.41 & \textbf{0.17}/\textbf{0.21}\\
Current + prior & 0.10/2.13 & 11.44/11.93 & 0.21/0.25 & \textbf{0.17}/\textbf{0.21}\\
Interval program & 0.22/2.26 & 2.26/6.17 & 0.43/0.48 & 0.39/0.43\\
Full archive & 0.22/2.26 & 2.26/6.17 & 0.43/0.48 & 0.39/0.43\\
Full belief & \textbf{0.02}/\textbf{2.04} & 2.72/3.87 & \textbf{0.03}/\textbf{0.07} & 0.80/0.85\\
\bottomrule\end{tabular}
\end{table}

\paragraph{Frozen state and observation protocol.} Grid indices $(a,b)\in\{0,\ldots,15\}^2$ represent $(a,b)/15$ with $8\times8$ coarse cells. Current $a\in\{5,6,9,10\}$, $b=a+\Delta$, and $\Delta\in\{-3,-1,1,3\}$. From $(b-2,a-4)$, three readbacks precede push-$A$ by $2$, push-$B$ by $4$, and exchange. Past images use canonical current $a=7$, recentered on the detected apple; all four later placements share their pixels/packet. Relative positions define a diagonal band without exposing translation. Unavailable readbacks publish $\varnothing$ while actions continue for every method; valid broad bands remain. Current views are full/upper-half, with missing coordinates unrestricted. The predeclared $12$-pixel tolerance is checked on $32$ separate calibration renders (seeds $90010,90011$); differences allow $24$ pixels. Filter history by $C=\pi^{-1}(\pi(S))$, but outputs/oracles use raw $S$.

Preserving requests use push-$A$ by $2$, exchange, then push-$B$ by $-2$; erasing requests use $x_B\leftarrow x_A$, push-$A$ by $2$, then exchange. Score only the three resulting future frames. SD1.5/Canny ControlNet/LCM-LoRA uses public flat-color two-shelf initialization, strength $0.8$, eight input/six effective steps, guidance $1.5$, and control $0.8$. The neural study crosses all $32$ seeds; confidence intervals resample whole seeds with all conditions.

\paragraph{Controls and scoring.} RT-DETR uses confidence $0.3$ and deduplication IoU $0.5$. Positions use a dominant expected-colour median inside a box (at least $16$ pixels), otherwise its centre. Missing, multiple, or wrong-track targets map to $\bot$. Both parsers project positions at most $12$ pixels beyond the public state range onto its endpoint; larger overshoots map to $\bot$. This fixed domain guard is independent of observation calibration. Whole-track HSV medians require $256$ pixels: PIL red hue $[0,15]\cup[240,255]$, blue $[140,185]$, saturation $\ge80$, value $\ge40$. Recursive beliefs predict then intersect available sets. Difference intervals number $1+31\cdot32/2=497$; action words number $\sum_{k=0}^3 3^k=40$. Charging all three interval-program slots and their count gives $\sum_{m=0}^3(497\cdot40)^m<2^{43}$ symbols, or $48$ byte-padded bits. Parsing and readout are target-blind.

\paragraph{Coverage and state recovery.} Membership is checked against the prescribed object-position command in every image-derived set. In each condition, current sets cover $512/512$ states; actual and corrected archives cover $504$, with eight empty-history fallbacks. Table~\ref{tab:coverage-check} gives complete set sizes and paired state-loss effects, using $20{,}000$ resamples of thirty-two whole seeds. All $2{,}016$ covered contexts satisfy $e_{\rm state}\le F^C+\mathcal E^C\le F^C+R_C$; all $2{,}048$ satisfy the fallback sound-set bound.

\begin{table}[t]
\centering\small\setlength{\tabcolsep}{3pt}
\caption{\textbf{Complete coverage and state-recovery analysis.} Each row contains all $512$ worlds. Current and sound sets cover $512$ states; the actual archive covers $504$, with eight empty-history fallbacks. $\Delta$ is current-only minus CAM physical state loss, scaled by $10^3$, with a paired whole-seed $95\%$ interval. Set sizes are means.}
\label{tab:coverage-check}
\begin{tabular}{@{}llrrl@{}}\toprule
View & Future & $|S|$ & $|S_{\rm sound}|$ & $10^3\Delta$ [interval]\\\midrule
Full & Preserve &1.39&1.10&1.181 [0.317,2.253]\\
Full & Erase &1.39&1.10&0.495 [$-0.651$,1.927]\\
Upper & Preserve &18.41&4.74&19.497 [18.863,20.052]\\
Upper & Erase &18.41&4.74&0.486 [$-0.234$,1.710]\\
\bottomrule
\end{tabular}
\end{table}

\paragraph{Observation calibration and rendering fidelity.} The independent development fit has $62/64$ valid object locations and $27$ valid relative measurements. Their largest global and half-relative residuals are $7.448$ and $8.659$ pixels, below the frozen $12$-pixel observation tolerance. Table~\ref{tab:renderer-calibration} reports RT-DETR and HSV control residuals for every development and future image, including invalid outputs' penalties.

\begin{table}[t]
\centering\small\setlength{\tabcolsep}{3pt}
\caption{\textbf{Independent renderer calibration and complete test transfer.} Parsed-to-command RMS uses every development image, including failures. $\widehat\varepsilon$ is the development maximum; coverage counts all $2,950$ future images below it.}
\label{tab:renderer-calibration}
\begin{tabular}{@{}lrrrl@{}}\toprule
Metric & Cal. invalid & $\widehat\varepsilon$ & Test invalid & Test coverage\\\midrule
RT-DETR &2/32&0.707110&192/2950&97.119\%\\
HSV &\textbf{0/32}&\textbf{0.016369}&\textbf{8/2950}&\textbf{99.153\%}\\\bottomrule
\end{tabular}
\end{table}

Valid RT-DETR development images have residual at most $0.016369$; retaining all test images at this tighter threshold gives $93.458\%$ coverage. Development maxima provide empirical transfer checks. On upper-preserve queries, the direct sound-set envelope at physical readout's output decreases from mean $0.24059$ using $S$ to $0.01324$ using $S\cap Q$. 

\paragraph{Excess and outer readout on one packet.} The two outputs agree in $2{,}040/2{,}048$ queries. Full-view decisions are identical. Upper-preserve has four changes, with mean excess/state/image differences $(-0.0617,0.0955,0.1010)\times10^{-3}$; upper-erase has four, with $(0.4762,-0.1562,-0.1595)\times10^{-3}$. All changed readbacks are valid under both parsers. This isolates state choice from invalid generation for these eight changed decisions.

\paragraph{Image readbacks and thresholds.} Upper-preserve RT invalid-readout counts are $68/512$ for CAM excess and $35/512$ for current minimax, with unit-failure floors $0.06934/0.03418$. At a diagnostic $1.1$-step threshold, CAM excess/full-archive accuracy is $82.2/80.7\%$. The threshold sweep assigns invalid outputs their prescribed penalties.

\paragraph{Physical readout and matched sound-pool control.} We use disagreement bound $f=1$ and output domain $S$. Physical readout and sound-pool minimax use the identical truth pool $S\cap Q_C$ and the same primary worst-truth objective. The former uses exact historical excess, then mean pool loss and state index, to resolve physical ties; the latter uses only the last two rules. Their outputs differ on seven of all $2{,}048$ contexts. Physical certificates agree in every context, and the exact secondary objective never increases the excess certificate. Actual image differences have both signs across the four conditions (Table~\ref{tab:neural-all}). This separates historical tie resolution from joint conditioning. All fifteen methods share the frozen renderer; every selected sequence is evaluated from its actual images.

\paragraph{Profile and physical-readout cost.} The single-thread benchmark uses all $2{,}048$ contexts, one excluded warm-up and five alternating paired passes. Family construction, conditioning, the full oracle profile and the final decision are included; perception, table construction, file access and rendering are excluded. Median pass means are $0.9545/0.9811$ ms for excess/physical readout. The shared table occupies $524{,}288$ bytes; peak traced working allocations are $332{,}973/332{,}727$ bytes, distinct from process and model memory. Generation takes $4.22$ seconds per frame with six effective denoising steps.

\subsection{Observation tolerance and raw-set stability}
The $6,9,12,15,18$-pixel scan retains the fixed generator, detector, metric guard and primary protocol: all $133{,}120$ sequences are scored with $20{,}000$ whole-seed resamples. Coverage and empty-history counts are unchanged. The continuous raw-set margin is nearest opposite-membership distance within $X$, minimized over the nonempty archive. Testing all four fine states per historical cell gives raw-$S$ constancy on $0\%$ of full-view and $0.99\%$ of upper-view archives. Historical filtering uses the cellwise-constant coarse closure $C$, for which finite conditioning remains exact. 

\clearpage
% Supplementary bibliography: original entries, inlined.
\end{document}